\documentclass[twoside,lettersize,journal]{IEEEtran}
\usepackage{amsmath,amsfonts}
\usepackage[ruled,vlined]{algorithm2e}
\usepackage{array}
\usepackage[caption=false,font=normalsize,labelfont=sf,textfont=sf]{subfig}
\usepackage{textcomp}
\usepackage{stfloats}
\usepackage{url}
\usepackage{verbatim}
\usepackage{graphicx}
\usepackage{cite}
\usepackage{orcidlink}
\usepackage[numbers]{natbib}
\usepackage{amsmath} 
\usepackage{bm}      
\usepackage{bbm}
\usepackage{multirow}
\usepackage{color} 
\usepackage{arydshln}
\newcommand{\first}[1]{\textcolor{red}{\bm{#1}}}
\newcommand{\second}[1]{\textcolor{blue}{\bm{#1}}}
\begin{document}

\title{LH$^2$Face: Loss function for Hard High-quality Face}

\author{
	Fan Xie$^{1, 2}$$^{~\orcidlink{0009-0006-6674-5103}}$,
	Yang Wang$^{1, 3,  \ddag}$$^{~\orcidlink{0000-0001-5386-7888}}$,~\IEEEmembership{Member,~IEEE,}\\
	Yikang Jiao$^{1}$$^{~\orcidlink{0009-0001-7316-4431}}$,
	Zhenyu Yuan$^{1}$$^{~\orcidlink{0009-0008-7013-3323}}$,
	Congxi Chen$^{1}$$^{~\orcidlink{0009-0000-1365-8280}}$,
	Chuanxin Zhao$^{1}$$^{~\orcidlink{0000-0002-5863-0430}}$,~\IEEEmembership{Member,~IEEE}\\
	$^{1}$Anhui Normal University, $^{2}$Chinese Academy of Agricultural Sciences\\
	\tt\small{ahwycap@mail.ahnu.edu.cn}, \tt\small{xiefan238@gmail.com}
	\thanks{$\ddag$ corresponding author.}
}


\markboth{}%
{}

\IEEEpubid{}

\maketitle

\begin{abstract}
In current practical face authentication systems, most face recognition (FR) algorithms are based on cosine similarity with softmax classification. Despite its  reliable classification performance, this method struggles with hard samples. A popular strategy to improve FR performance is incorporating angular or cosine margins. However, it does not take face quality or recognition hardness into account, simply increasing the margin value and thus causing an overly uniform training strategy.
To address this problem, a novel loss function is proposed, named Loss function for Hard High-quality Face (LH$^2$Face).
Firstly, a similarity measure based on the von Mises-Fisher (vMF) distribution is stated, specifically focusing on the logarithm of the Probability Density Function (PDF), which represents the distance between a probability distribution and a vector. Then, an adaptive margin-based multi-classification method using softmax, called the Uncertainty-Aware Margin Function, is implemented in the article. Furthermore, proxy-based loss functions are used to apply extra constraints between the proxy and sample to optimize their representation space distribution. Finally, a renderer is constructed that optimizes FR through face reconstruction and vice versa.
Our LH$^2$Face is superior to similiar schemes on hard high-quality face datasets, achieving 49.39\% accuracy on the IJB-B dataset, which surpasses the second-place method by 2.37\%.
\end{abstract}


\section{Introduction}
\IEEEPARstart{F}{ace} recognition (FR) is a widely studied field in information security over the past ten years. This technology is applied in areas such as electronic payments, account logins, and public transportation access authentication. It remains a field of great attention and importance, which attracts many researchers work on it.

Various kinds of FR systems have been raised. Metric learning~\cite{contrastiveloss, facenet} was one of the earliest algorithms introduced for FR, based on sample-to-sample comparison. Although good results were yielded, it faced challenges with convergence and had significant data mining difficulties. This was followed by the dominance of cosine similarity algorithms~\cite{liu2017sphereface, wang2018cosface, deng2019arcface} leading to the era of proxy-sample and relative distance (RD) methods. RD performs as the difference between the similarity of the proxy to the positive sample and negative sample. After that, many efficient algorithms based on ArcFace~\cite{deng2019arcface} have been proposed~\cite{kim2022adaface, meng2021magface, huang2020curricularface} to improve the accuracy of FR systems. Meanwhile, some algorithms integrating face generation~\cite{3d-berl, dcface} have also contributed to the development of the FR field.

\begin{figure}[h]
	\begin{center}
		\includegraphics[width=1.0\linewidth]{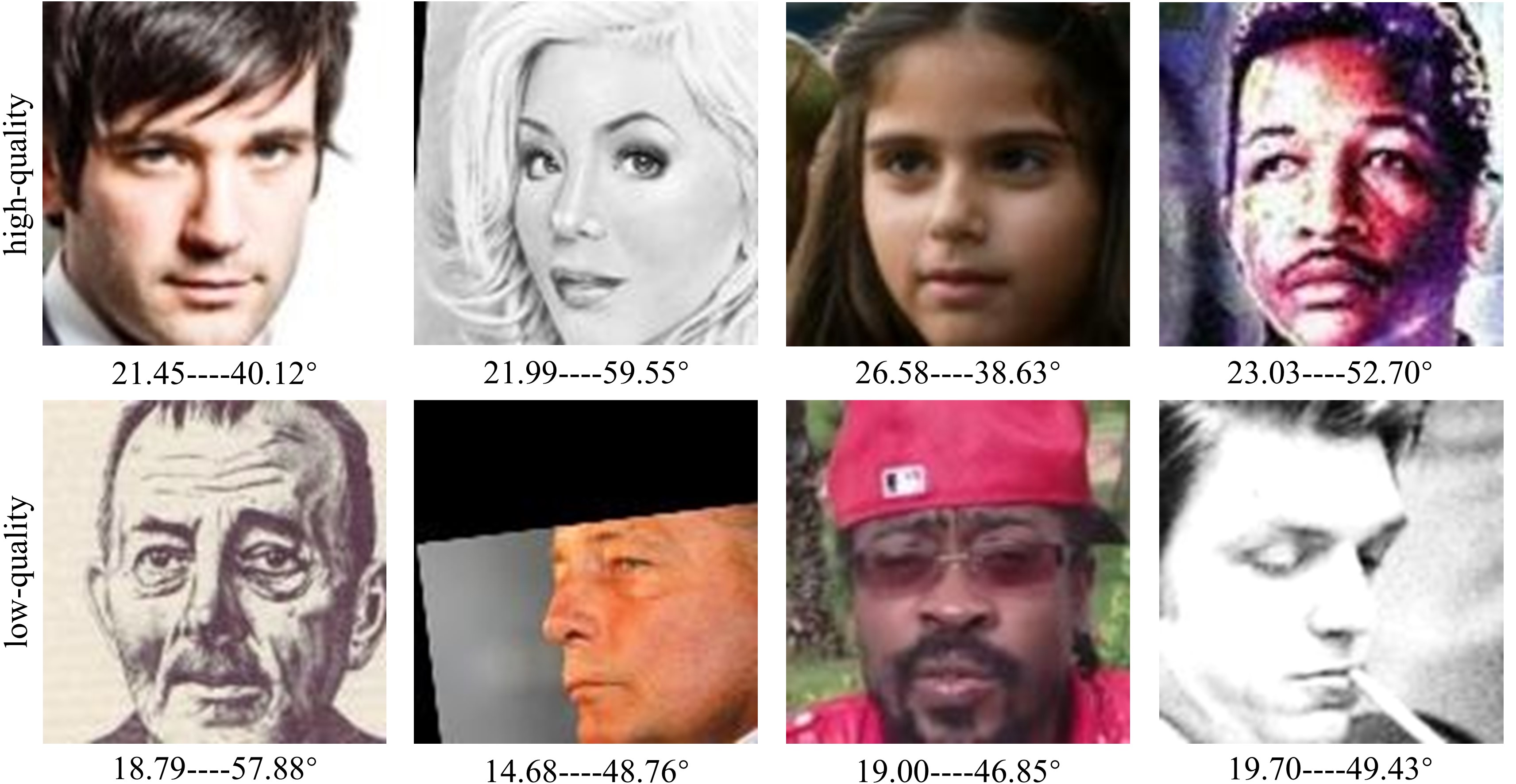}
	\end{center}
	\caption{The first line in the image shows high-quality samples, while the second line shows low-quality samples. The numbers below each image represent the feature norm (the L2 norm of the feature vector) and the angle between the image’s feature vector and the corresponding proxy feature vector. For example, $21.45-40.12^{\circ}$ means the feature norm is $21.45$, and the angle is $40.12^{\circ}$. It can be observed that a relatively simple classification method has been used here, with a feature norm of $20$ as the boundary between high-quality and low-quality samples. The value of this boundary is not strictly accurate, or rather, it cannot be perfectly quantified. The number $20$ is used here simply as a boundary for the sake of convenience in understanding. The larger the angle between the image's feature vector and the corresponding proxy feature vector, the harder the sample is.}
	\label{face}
\end{figure}

\IEEEpubidadjcol

In recent years, many algorithms have focused on recognizing low-quality images in Fig.~\ref{face} (such as those with excessive blur and significant occlusion). In contrast, this paper takes a different route. Low-quality images  easily leads to semantic ambiguity and consequently a clear upper limit on recognition accuracy because of lacking significant information. Due to this reason, it is argued that improvements in performance on low-quality datasets do not necessarily reflect the true recognition capability of an algorithm. Such improvements are only considered accurate if the algorithm includes deblurring techniques during recognition. Compared to low-quality datasets, hard high-quality images (with large poses and slight blur or occlusion) pose a greater challenge. These images, containing enough information but remain hard to recognize, highlight the current imperfections of FR algorithms.

Additionally, methods~\cite{liu2017sphereface, wang2018cosface, deng2019arcface, kim2022adaface, meng2021magface, huang2020curricularface} based on cosine complexity ignore the absolute distance (AD) during recognition. AD is the distance between the proxy and the positive sample, as well as between the proxy and the negative sample. The former is referred as positive absolute distance (PAD), and the latter is negative absolute distance (NAD). These approaches hinder the deep optimization of the proxy vector space.

The goal of our research is to improve the recognition accuracy on hard high-quality datasets. To achieve it, high-quality face images are distinguished from low-quality ones by using feature norms. Based on margin and the von Mises-Fisher (vMF) distribution, a new similarity measurement method is proposed. This method assigns smaller concentration values to low-quality images to minimize their interference on the recognition of high-quality images. A proxy-based loss is also introduced to optimize the AD. Finally, a renderer is constructed to operate face reconstruction, which assists in enhancing FR performance. During the training process, the performance of face reconstruction also shows notable improvement.

In summary, the main contributions of our work are as follows: 
\begin{itemize}
	\item A loss function based on margin and the vMF distribution is proposed to differentiate face images of varying quality. In this framework, the similarity measure, which is an essential component of the loss function, is implemented adaptively through the feature norm.
	\item Three proxy-based loss functions are presented to optimize the AD, and as a result, the proxy's feature space is also improved.
	\item Several loss functions are stated to optimize the performance of face reconstruction and enhance FR through face reconstruction, with a renderer constructed for depth map rotation in 3D reconstruction.
\end{itemize}

\section{Related Work}
\subsection{Face Recognition} 
In metric learning, the most widely used methods are contrastive loss~\cite{contrastiveloss} and triplet loss~\cite{facenet}, both relying on margin. These methods provide absolute margin, while softmax and margin-based methods offer relative margin. Many absolute margin variants based on contrastive loss and triplet loss, such as lifted structure loss~\cite{Lifted_Structure_Loss} and binomial deviance loss~\cite{Binomial_Deviance_Loss}, also have an inherent connection with recent pairwise methods. Clearly, margin-based methods have been widely applied in FR.

Cross-entropy loss is commonly used for image classification. When cross-entropy is combined with cosine similarity~\cite{liu2017sphereface, wang2017normface}, it better aligns with the characteristics of FR. Subsequently, researchers introduced margin~\cite{wang2018cosface, deng2019arcface} to reduce intra-class distance, further optimizing the performance of FR. These methods distinguish between positive and negative samples by ensuring that the maximum intra-class distance is smaller than the minimum inter-class distance.

MagFace~\cite{meng2021magface} introduced an adaptive mechanism that causes easier samples to move closer to the class center with an increased margin. It means harder samples are pushed farther away from the class center with a reduced margin, thus the intra-class feature distribution is optimized.
AdaFace~\cite{kim2022adaface} pioneered the application of feature norm in FR, proposing a feature-norm-based learning method to address the recognition problem of low-quality faces.
PFE~\cite{shi2019probabilistic} extended class representations from a single point to a Gaussian distribution, using expected mutual information to handle inter-class uncertainty, improving the robustness of FR.
SCF-ArcFace~\cite{li2021spherical} adopted a spherical confidence learning method, using concentration values as confidence measures and combining Kullback-Leibler (KL) divergence with the spherical Dirac $\delta$ function to enhance probabilistic distribution in FR.
FRABSM~\cite{FRABSM} improved FR performance by emphasizing the relative importance of hard samples versus easy samples.

DCFace~\cite{dcface} significantly improved FR accuracy by training models with images generated by a dual-condition face generator based on a diffusion model.
ReFO~\cite{refo} found that the distillation performance was poor without identity supervision, leading to the identification of the gap between intrinsic dimensions and capacity, ultimately proposing a reverse distillation method.
VLRFR~\cite{vlrfr2} discovered a bimodal distribution and pulled hard samples out from indistinguishable ones.
IIC~\cite{iic} applied knowledge distillation and introduced feature decomposition, cleverly enhancing FR performance.
KP-RPE~\cite{kprpe} optimized face alignment through relative position encoding, thereby improving FR performance.
LAFS~\cite{lafs} proposed landmark-based self-supervised learning to learn generalized representations that can be adapted for highly accurate FR.
QSD~\cite{qsd} introduced a quality-aware feature refinement framework based on a dedicated quality prior obtained from recognition performance.
CoReFace~\cite{coreface} adopted sample-guided contrastive learning, directly regularizing training based on sample-to-sample relationships, ensuring consistency with the evaluation process.
Joint-BERL~\cite{joint-berl} constructed a 3D reconstruction bypass and a blind restoration bypass, integrating self-supervised and supervised learning to aid in robust feature learning for FR.
FRL~\cite{frl} introduces a simple yet effective method that enhances FR across varying poses by using a representation generator to reconstruct face features, minimizing identity differences and incorporating pixel-level and adversarial constraints for improved model discrimination.
VirFace$^\infty$~\cite{virface} introduces a novel semi-supervised FR method that effectively utilizes unlabeled shallow data by applying virtual classes and virtual distributions to significantly enhance recognition performance.

\subsection{Face Reconstruction}
ID-GAN~\cite{id-gan} proposed a method based on an identity-diversity generative adversarial network, resulting in improving the performance of FR systems on occluded faces by reconstructing occluded faces. 
Unsup-3d~\cite{unsup3d} introduced a 3D model capable of learning a deformable object category, enabling high-fidelity monocular 3D reconstruction of individual object instances from an unconstrained single view. This approach trains without any supervision based on reconstruction loss, similar to an autoencoder.  
FaceCycle~\cite{face-cycle} decouples face expressions from identity representations by using cyclic consistency constraints on face motion and identity. Through unsupervised training, the model learns these two representations from large-scale face data and can be applied to various downstream tasks, such as expression recognition, head pose regression, identity recognition, and face image frontalization.  
LAP~\cite{lap} decomposes faces into identity-consistent geometry/texture and scene-specific details. Identity-consistent geometry/texture primarily represents the basic structure of the face, while scene-specific details (such as expression, lighting, etc.) are adjusted according to the input image. To handle diverse and uncertain real-world image datasets (e.g., face expressions, makeup, skin variations), LAP employs a curriculum learning strategy, starting from training with simple synthetic data and gradually transitioning to more complex and realistic outdoor image datasets.

MDFR~\cite{mdfr} achieves face frontalization and face restoration through 3DMM and adversarial learning. It requires multi-stage training with multiple datasets and is a supervised learning approach.  
3D-BERL~\cite{3d-berl} improves the backbone FR network by introducing a 3D face reconstruction loss with two auxiliary networks, allowing the recognition network to understand faces in 3D views.  
L2R~\cite{l2r} enables unsupervised high-quality face reconstruction from low-resolution images.  
PhyDIR~\cite{phydir} separates the "3D geometry" learning and "neural rendering" components into distinct training processes. The advantage of this approach is that the model first learns to recover accurate face geometry and then progressively refines the texture and rendering. By combining implicit textures with a physical graphics pipeline, it avoids some limitations of traditional methods.  
IGC-Net~\cite{igc-net} utilizes a one-hot attention mechanism and employs auxiliary losses such as contrastive loss, sparsity loss, and background separation loss to achieve image segmentation and face reconstruction.

\section{Proposed Method}
This section provides a detailed introduction to our method. It consists of three parts. The loss function in the first part serves as the main loss during training, with a very high weight, which consequently takes over the centrel point of this paper. Auxiliary loss functions in the second and third parts work to help optimize the training process. The proposed method is named Loss function for Hard High-quality Face (LH$^2$Face).
\subsection{Uncertainty-Aware Margin Function}
\subsubsection{von Mises-Fisher Distribution}
\label{}
The vMF distribution is employed as a tool to establish a new method for measuring similarity.
Specifically, for a given face image $ \bm{z} $ from the feature space $ \mathcal{X} $ (output of the neural network), the corresponding normalized vector $ \bm{x} $ is computed as $ \bm{x} = \frac{\bm{z}}{\|\bm{z}\|} $.
The distribution is modeled as a vMF distribution defined on the $ n $-dimensional unit sphere $ \mathbb{S}^{n-1} \subset \mathbb{R}^n $, and its form is expressed as follows:
\begin{equation}
	f_{p}(\bm{x}; \bm{z}) = \frac{\kappa_{\bm{z}}^{n/2-1}}{(2\pi)^{n/2}\mathcal{I}_{n/2-1}(\kappa_{\bm{z}})} \exp \left( \kappa_{\bm{z}} \bm{\mu}^\text{T} \bm{x} \right)
\end{equation}
The subscript indicates the dependence on $ \bm{z} $ that will be omitted later. Here, $ \bm{x}, \bm{\mu} \in \mathbb{S}^{n-1} $, $ \kappa \geq 0 $. The parameters $ \bm{\mu} $ and $ \kappa $ are called the mean direction and concentration parameter, respectively. The concentration parameter $ \kappa $ controls how tightly the distribution is centered around the mean direction $ \bm{\mu} $. When $ \kappa > 0 $, the distribution is unimodal; when $ \kappa = 0 $, the distribution degenerates to a uniform distribution on the sphere. $ \mathcal{I}_\alpha $ represents the modified Bessel function of the first kind of order $ \alpha $, defined as follows:
\begin{equation}
	\mathcal{I}_\alpha(x) = \sum_{m=0}^{\infty} \frac{1}{m! \Gamma(m+\alpha+1)} \left(\frac{x}{2}\right)^{2m+\alpha}
\end{equation}

\begin{figure}[h]
	\begin{center}
		\includegraphics[width=1.0\linewidth]{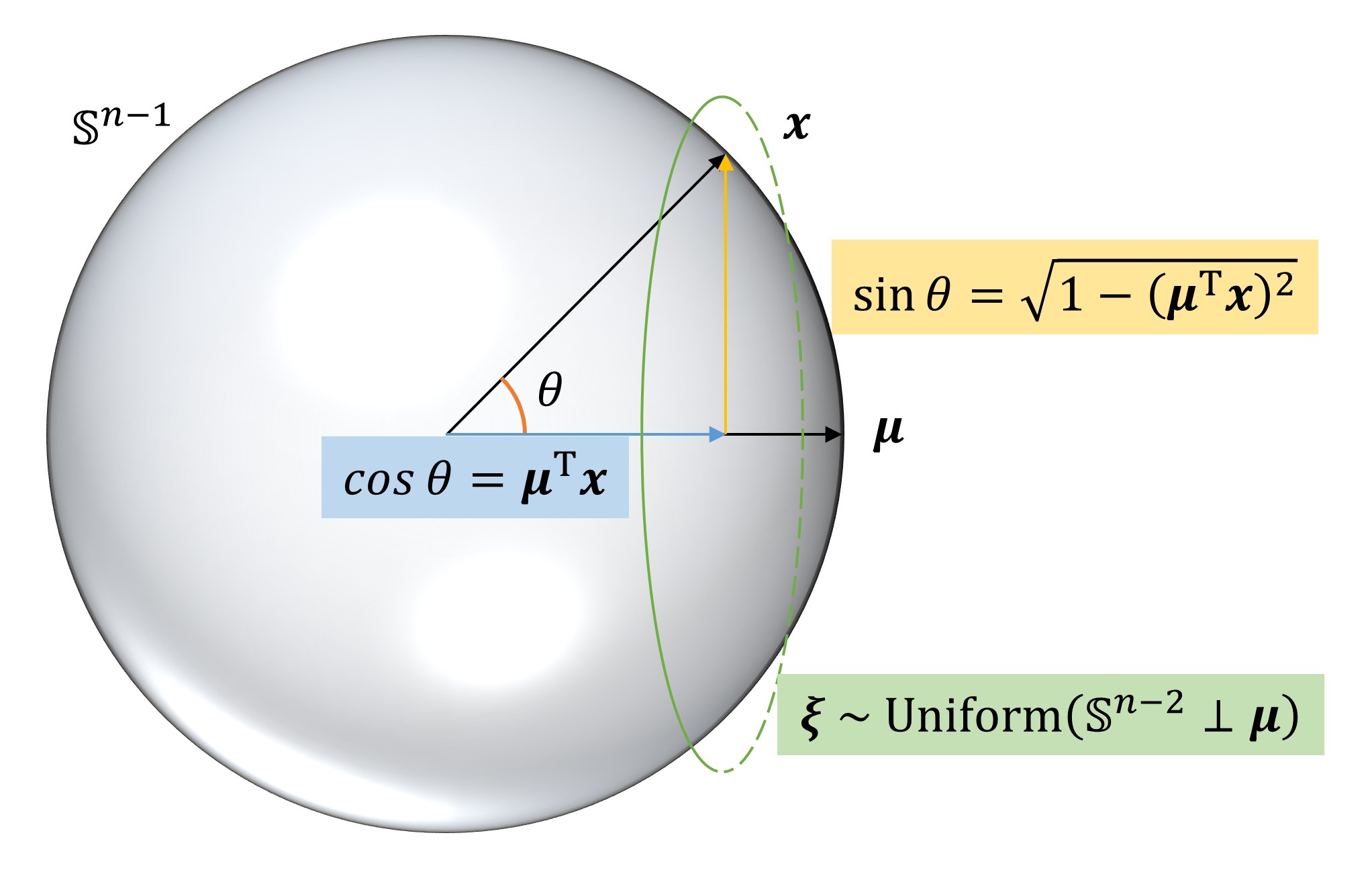}
	\end{center}
	\caption{This is a visualization of the values of $ \bm{\mu}^\text{T} \bm{x} $ on the unit sphere, where $ \bm{\xi} $ is a vector on the sphere that satisfies $ \cos\theta = \bm{\mu}^\text{T} \bm{x} $. It belongs to a ring of dimension $ n - 2 $.}
	\label{figure_ball}
\end{figure}

\begin{figure}[h]
	\begin{center}
		\includegraphics[width=1.0\linewidth]{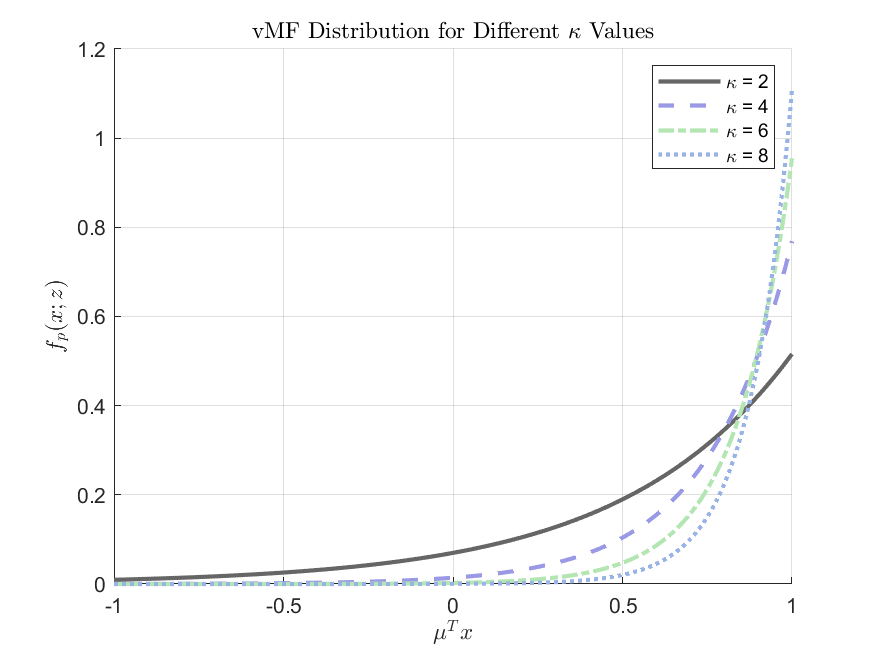}
	\end{center}
	\caption{This is the plot of the vMF distribution for $ n = 2 $, showing the relationship between the PDF and $ \bm{\mu}^\text{T} \bm{x} \in \left[-1, 1\right] $.}
	\label{vmf_distribution_cosine_similarity}
\end{figure}

From Figs.~\ref{figure_ball} and ~\ref{vmf_distribution_cosine_similarity}, it can be observed that the horizontal axis $ \bm{\mu}^\text{T} \bm{x} $ is equivalent to the cosine similarity. 
It can be observed that the logarithm of the Probability Density Function (PDF) can be used as a substitute for the cosine similarity. It provides sufficient discriminative power and can scale the similarity based on the value of $ \kappa $.

\subsubsection{Feature Dimension}
\label{}
When $ n = 512 $ and $ \kappa < 14 $, the value of $ \mathcal{I}_{n / 2 - 1}(\kappa) $ becomes $ -\infty $, meaning it cannot be computed. Therefore, $n$ is forced to be set to 256 or 128, which leads us to consider whether it is reasonable for the dimensionality of the feature vector $ d $ to be inconsistent with $ n $.

After analysis, the vMF distribution retains its distributional characteristics across different dimensions $ n $. Specifically:
\begin{itemize}
	\item The vMF distribution is a probability distribution defined on high-dimensional spheres, used for modeling directional data. 
	\item The dimension $ n $ only affects the shape of the distribution and the normalization constant, but does not alter the fundamental characteristics of the distribution. 
	\item Changing the dimension from $ 512 $ to another still preserves the characteristic of the data being distributed on the sphere.
\end{itemize}
Therefore, from a probabilistic perspective, the definition of the vMF distribution remains valid. Of course, similar to UCFace~\cite {ucface}, adding an autoencoder layer to adjust the value of $ d $ is one way to achieve $ d = n $. However, this approach was not adopted. Instead, $d$ was allowed to be different from $n$ directly. In fact, if an autoencoder layer is used, the approach becomes a variant of UCFace.

\subsubsection{Margin Function}
\label{}

\begin{figure*}[h]
	\begin{center}
		\includegraphics[width=1.0\linewidth]{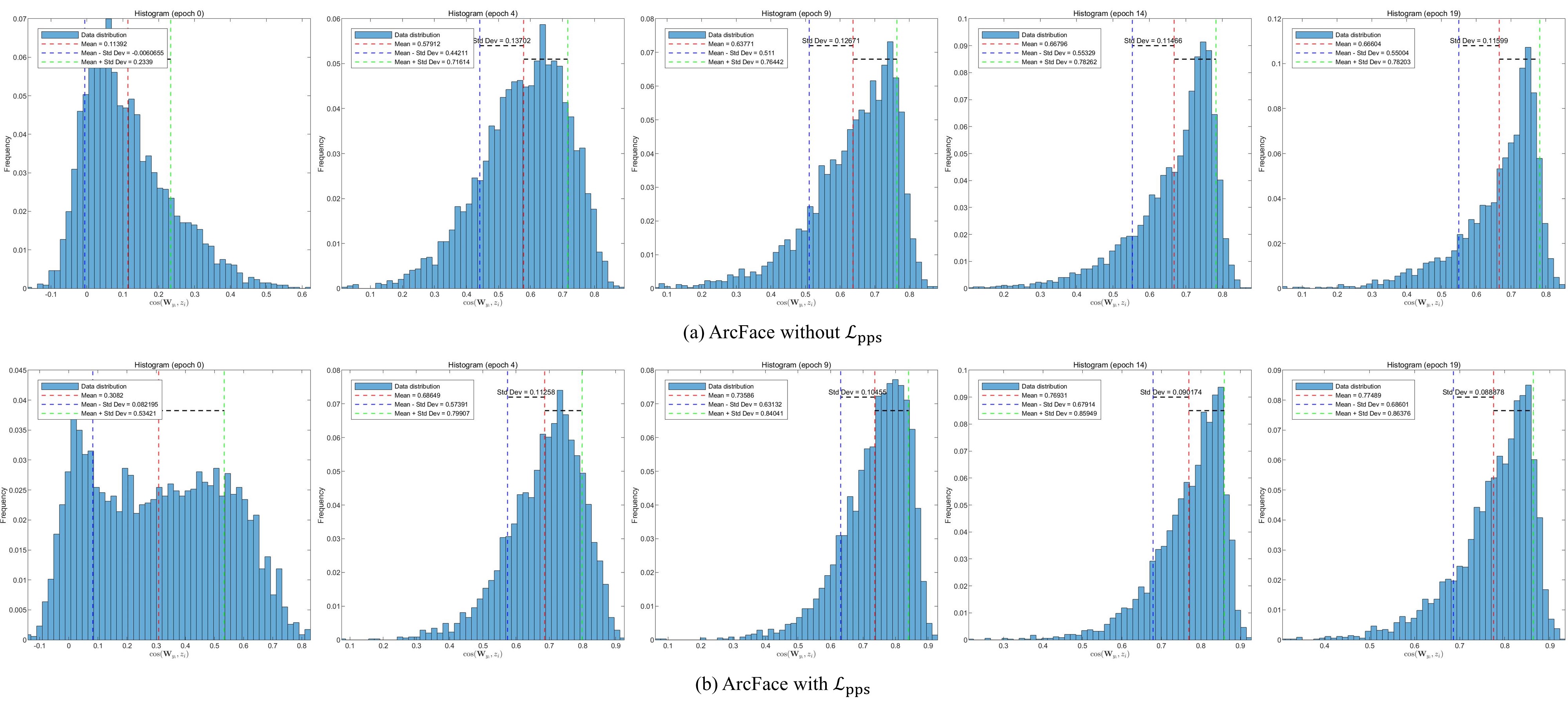}
	\end{center}
	\caption{The image shows the similarity distribution between the proxies of ArcFace before and after adding $\mathcal{L}_\text{pps}$, along with their corresponding positive samples.}
	\label{figure_cos_histogram}
\end{figure*}

\begin{algorithm}[t]
	\SetAlgoLined \caption{\small the Uncertainty-Aware Margin Function}
	\label{algorithm1}
	\begin{small}
		\KwIn{The hyper-parameter $\tau = 1.0$, $\mu_{\|\bm{z}\|}^{(0)} = 20$, $k = 1$.}
		
		\KwOut{Parameters $\bm{\Theta}$ and $\mathbf{W}$.}
		
		1: \textbf{for $e=1:eps$ do}
		
		2:\quad \textbf{for $b=1:n$ do}
		
		3:\qquad Calculate $ \mu_{\|\bm{z}\|} = \alpha \mu_{\|\bm{z}\|}^{(k)} + (1 - \alpha) \mu_{\|\bm{z}\|}^{(k-1)} $ and $m = 0.35 \mu_{\|\bm{z}\|}$;
		
		4:\qquad \textbf{for $i=1:N$ do}
		
		5:\qquad\quad Get $\mathcal{L}_{\text{vMF}}^i$ by Eqs.~\ref{vMF-Loss} and \ref{h-eq};
		
		6:\qquad \textbf{end for}
		
		7:\qquad Calculate $\mathcal{L}_\text{vMF}$ within the current mini-batch by Eq.~\ref{vMF-Loss};
		
		8:\qquad Calculate the gradient values for the current mini-batch by
		$\frac{\partial \mathcal{L}_\text{vMF}}{\partial \mathbf{W}^{(b)}}=\sum_{i=1}^{N}\frac{\partial \mathcal{L}_\text{vMF}}{\partial \mathbf{W}_{i}}$, $\frac{\partial \mathcal{L}_\text{vMF}}{\partial \bm{x}^{(b)}}=\sum_{k=1}^{N}\frac{\partial \mathcal{L}_\text{vMF}}{\partial \bm{x}_y^{(k)}}$;
		
		9:\qquad Update the parameters $\mathbf{W}$ and $\bm{\Theta}$ by
		$\mathbf{W}^{(b+1)}  =\mathbf{W}^{(b)}-\lambda^{(b)} \frac{\partial \mathcal{L}_\text{vMF}}{\partial \mathbf{W}^{(b)}}$,
		$\boldsymbol{\Theta}^{(b+1)}  =\boldsymbol{\Theta}^{(b)}-\lambda^{(b)} \frac{\partial \mathcal{L}_\text{vMF}}{\partial \bm{x}^{(b)}} \frac{\partial \bm{x}^{(b)}}{\partial \bm{\Theta}^{(b)}}$;
		
		10:\quad \textbf{end for}
		
		11: \textbf{end for}
		
		12: Return CNN parameter $\bm{\Theta}$ and last FC parameter $\mathbf{W}$.
	\end{small}
\end{algorithm}

The value of $ \kappa $ is directly set to $ |\bm{z}| $. This approach allows the quality of the image to be directly related to the variance of the distribution. At this point, the similarity in the feature space has been established, that is:
\begin{equation}
	\begin{split}
		&\text{sim}\left(\mathbf{W}_j, \bm{z}_i\right) = \log f_{p}(\bm{x}; \bm{\mu} = \mathbf{W}_j, \bm{z} = \bm{z}_i)\\
		=&\log \left(\frac{\kappa^{n / 2-1} \exp \left(\kappa \bm{\mu}^{\text{T}} \bm{x}\right)}{(2 \pi)^{n / 2} \mathcal{I}_{n / 2-1}\left(\kappa\right)}\right) \\
		=&\kappa \cos \theta+(n / 2-1) \log \kappa-(n / 2) \log (2 \pi)\\
		&-\log \left(\mathcal{I}_{n / 2-1}\left(\kappa\right)\right)
	\end{split}
\end{equation}
where $\mathbf{W}$ means the proxies. $\text{sim}\left(\mathbf{W}_j, \bm{z}_i\right)$ represents the similarity between the feature vector of the $ j $-th class and the $ i $-th image. Based on this similarity, the Uncertainty-Aware Margin Function (UAMF) can be derived as follows:
\begin{equation}
	\begin{split}
		&\mathcal{L}_{\text{vMF}}=\frac{1}{N} \sum_{i=1}^{N} \mathcal{L}^{i}_{\text{vMF}}, \\
		&\mathcal{L}^{i}_{\text{vMF}} = -\log \frac{\exp \left(h\left(\mathbf{W}_j, \bm{z}_i, m, \tau\right)\right)}{\sum_{j=1}^{C} \exp \left(h\left(\mathbf{W}_j, \bm{z}_i, m, \tau\right)\right)}
	\end{split}
	\label{vMF-Loss}
\end{equation}
and
\begin{small}
	\begin{equation}
		h\left(\mathbf{W}_j, \bm{z}_i, m, \tau\right)=\left\{\begin{array}{cc}
			\left(\text{sim}\left(\mathbf{W}_j, \bm{z}_i\right)-m\right)/\tau & j=y_i \\
			\text{sim}\left(\mathbf{W}_j, \bm{z}_i\right)/\tau & j \neq y_i
		\end{array}\right.
		\label{h-eq}
	\end{equation}
\end{small}
where $ y_i $ denotes the class label of the $ i $-th sample, corresponding to the $ y_i $-th column of $ \mathbf{W} $. $C$ is the total number of classes. $\tau$ is a scalar temperature parameter. For the margin value, it is set to $ m = 0.35 \mu_{|\bm{z}|} $, where $ \mu_{|\bm{z}|} $ is the mean of $ |\bm{z}| $ obtained from the EMA equation $ \mu_{|\bm{z}|} = \alpha \mu_{|\bm{z}|}^{(k)} + (1 - \alpha) \mu_{|\bm{z}|}^{(k-1)} $. Algorithm~\ref{algorithm1} demonstrate the training procedure for UAMF.

\subsection{Proxy-based Loss}
\label{}

PAD changed as the training progressed. The PAD values were tracked throughout each epoch. Fig.~\ref{figure_cos_histogram} shows the distribution of the PAD as training progresses. From observing Fig.~\ref{figure_cos_histogram} (a), it can be seen that as training progresses, the similarity between the proxy and the positive sample gradually increases. It stabilizes once reaching to a certain threshold, determined by the size of the margin. This is because most of the popular methods at present are based on RD. No penalty is applied when the RD exceeds a certain margin value:
\begin{equation}
	\text{if} \cos \left(\mathbf{W}_{y_i}, \bm{z}_i\right) - \cos \left(\mathbf{W}_j, \bm{z}_i\right) > m \text{ and } j \neq y_i, \mathcal{L}_{\text{FR}}^i \approx 0
\end{equation}
where $ \mathcal{L}_{\text{FR}} $ is the loss for FR. Before the introduction of triplet loss, RD was not an obvious approach. For example, contrastive loss would penalize the AD. However, after the introduction of triplet loss, all methods based on cosine similarity have ignored the AD. It must be acknowledged that excessive constraints on the AD can lead to overfitting. Nevertheless, when the weight of this constraint was set to a smaller value, positive improvements were observed in the experimental results.

\begin{algorithm}[t]
	\SetAlgoLined \caption{\small Proxy-based Loss}
	\label{algorithm2}
	\begin{small}
		\KwIn{The hyper-parameter $\cos \left(\mathbf{W}_{y}, \bm{z}\right)_{\text{min}} = 0.5$, $\cos \left(\mathbf{W}_{y}, \bm{z}\right)_{\text{max}} = 0.9$, $\lambda_\text{pps} = 10$, $\lambda_\text{pns} = 20$, $\lambda_\text{pp} = 150$, $N_{\text{left}} = 0$.}
		
		\KwOut{Parameters $\bm{\Theta}$ and $\mathbf{W}$.}
		
		1: Initialize $\left.\cos \left(\mathbf{W}_{y_i}, \bm{z}_i\right)_{\text{mid}}\right|_{0} = \cos \left(\mathbf{W}_{y}, \bm{z}\right)_{\text{min}}$;
		
		2: \textbf{for $e=1:eps$ do}
		
		3:\quad \textbf{for $b=1:n$ do}
		
		4:\qquad Retrieve $N_{\text{unique}}$ from the label;
		
		5:\qquad Calculate $\mathcal{L}_\text{pp}$ by Eq.~\ref{pp-loss};
		
		6:\qquad Calculate $\cos \left(\mathbf{W}_{y_0}, \bm{z}_0\right)$ for the first sample, denoted as $\cos \left(\mathbf{W}_{y}, \bm{z}\right)_b$, for use in calculating $\left.\cos \left(\mathbf{W}_{y_i}, \bm{z}_i\right)_{\text{mid}}\right|_{e}$ later;
		
		7:\qquad \textbf{for $i=1:N$ do}
		
		8:\qquad\quad \textbf{if} $\cos \left(\mathbf{W}_{y_i}, \bm{z}_i\right)<\left.\cos \left(\mathbf{W}_{y_i}, \bm{z}_i\right)_{\text{mid}}\right|_{e-1}$
		
		9:\qquad\qquad $ N_{\text{left}} = N_{\text{left}} + 1 $;
		
		10:\qquad \textbf{end for}
		
		11:\qquad Calculate $\mathcal{L}_\text{pps},  \mathcal{L}_\text{pns}$ using Eqs.~\ref{pps-loss} and \ref{pns-loss};
		
		12:\qquad Obtain $\mathcal{L}_\text{proxy-based}$ by Eq.~\ref{proxy-based loss};
		
		13:\qquad Reset $N_{\text{left}}$ to $0$;
		
		14:\qquad Calculate the gradient values for the current mini-batch by
		$\frac{\partial \mathcal{L}_\text{proxy-based}}{\partial \mathbf{W}^{(b)}}=\sum_{i=1}^{N}\frac{\partial \mathcal{L}_\text{proxy-based}}{\partial \mathbf{W}_{i}}$, $\frac{\partial \mathcal{L}_\text{proxy-based}}{\partial \bm{x}^{(b)}}=\sum_{k=1}^{N}\frac{\partial \mathcal{L}_\text{proxy-based}}{\partial \bm{x}_y^{(k)}}$;
		
		15:\qquad Update the parameters $\mathbf{W}$ and $\bm{\Theta}$ by
		$\mathbf{W}^{(b+1)}  =\mathbf{W}^{(b)}-\lambda^{(b)} \frac{\partial \mathcal{L}_\text{proxy-based}}{\partial \mathbf{W}^{(b)}}$,
		$\boldsymbol{\Theta}^{(b+1)}  =\boldsymbol{\Theta}^{(b)}-\lambda^{(b)} \frac{\partial \mathcal{L}_\text{proxy-based}}{\partial \bm{x}^{(b)}} \frac{\partial \bm{x}^{(b)}}{\partial \bm{\Theta}^{(b)}}$;
		
		16:\quad \textbf{end for}
		
		17:\quad Calculate $\left.\cos \left(\mathbf{W}_{y_i}, \bm{z}_i\right)_{\text{mid}}\right|_{e} = \frac{1}{n} \sum_{i=1}^{n} \cos \left(\mathbf{W}_{y}, \bm{z}\right)_i$;
		
		18:\quad \textbf{if} $\left.\cos \left(\mathbf{W}_{y_i}, \bm{z}_i\right)_{\text{mid}}\right|_{e} < \cos \left(\mathbf{W}_{y}, \bm{z}\right)_{\text{min}}$
		
		19:\qquad $\left.\cos \left(\mathbf{W}_{y_i}, \bm{z}_i\right)_{\text{mid}}\right|_{e} = \cos \left(\mathbf{W}_{y}, \bm{z}\right)_{\text{min}}$;
		
		20:\quad \textbf{elif} $\left.\cos \left(\mathbf{W}_{y_i}, \bm{z}_i\right)_{\text{mid}}\right|_{e} > \cos \left(\mathbf{W}_{y}, \bm{z}\right)_{\text{max}}$
		
		21:\qquad $\left.\cos \left(\mathbf{W}_{y_i}, \bm{z}_i\right)_{\text{mid}}\right|_{e} = \cos \left(\mathbf{W}_{y}, \bm{z}\right)_{\text{max}}$;
		
		22: \textbf{end for}
		
		23: Return CNN parameter $\mathbf{\Theta}$ and last FC parameter $\bm{W}$.
	\end{small}
\end{algorithm}

It can be observed that the PAD follows a skewed Gaussian distribution, while the NAD follows a standard Gaussian distribution. This distribution represents the degree of similarity between the sample and the proxy. When ResNet18 is replaced with ResNet50 or ResNet100, the mean of the PAD distribution increases, and the recognition performance improves as well.

The above analysis is qualitative and does not yet translate into a specific constraint method.
For the specific constraint methods, the following approach was adopted, proposing three proxy-based losses. The first is $ \mathcal{L}_\text{pps} $ which stands for the constraint on the PAD.
\begin{equation}
	\mathcal{L}_\text{pps}= \lambda_\text{pps} \frac{\sum_{i=1}^{N_{\text {left }}}\left(\cos \left(\mathbf{W}_{y_i}, \bm{z}_i\right)-\left.\cos \left(\mathbf{W}_{y_i}, \bm{z}_i\right)_{\text {mid }}\right|_{e-1}\right)^{2}}{N_{\text {left }}}
	\label{pps-loss}
\end{equation}
where $ \left.\cos \left(\mathbf{W}_{y_i}, \bm{z}_i\right)_{\text{mid}}\right|_{e-1} $ represents the clipped mean of the PAD from the previous epoch, with a minimum value of $\cos \left(\mathbf{W}_{y}, \bm{z}\right)_{\text{min}} = 0.5$ and a maximum value of $\cos \left(\mathbf{W}_{y}, \bm{z}\right)_{\text{max}} = 0.9$. $ N_{\text{left}} $ is the number of positive samples whose PAD are smaller than $ \left.\cos \left(\mathbf{W}_{y_i}, \bm{z}_i\right)_{\text{mid}}\right|_{e-1} $, and "pps" is the abbreviation for Proxy and Positive Sample. In other words, only approximately half of the samples are penalized to prevent overfitting. Our approach not only increases the PAD's mean value but also reduces the PAD's variance, achieving a twofold benefit.

For the NAD, the following constraint is also applied:
\begin{equation}
	\mathcal{L}_\text{pns}= \lambda_\text{pns} \frac{\sum_{i=1, j \neq y_{i}}^{N C}\cos^2 \left(\mathbf{W}_{j}, \bm{z}_i\right)}{N(C-1)}
	\label{pns-loss}
\end{equation}
where "pns" is the abbreviation for Proxy and Negative Sample.

In addition, making the proxies mutually orthogonal beacuse orthogonality indicates independence in feature space.
\begin{equation}
	\mathcal{L}_\text{pp}= \lambda_\text{pp} \frac{\sum_{i=1}^{\left(N+N_{\text {unique }}\right)\left(N+N_{\text {unique }}-1\right) / 2} \cos^2 \left(\mathbf{W}_{y_i}, \mathbf{W}_j\right)}{\left(N+N_{\text {unique }}\right)\left(N+N_{\text {unique }}-1\right) / 2}
	\label{pp-loss}
\end{equation}
where $ N $ is the batch size, and $ N_{\text{unique}} $ represents the number of unique proxies in the current batch size. The term "unique" is used because there may exist multiple positive samples for some proxies within a batch, so $ N_{\text{unique}} \leq N $. In this case, $ N $ proxies are randomly sampled, along with the $ N_{\text{unique}} $ proxies corresponding to the indices of the current batch size.

The final FR loss function is:
\begin{equation}
	\mathcal{L}_{\text{FR}} = \mathcal{L}_{\text{vMF}} + \mathcal{L}_\text{proxy-based}
	\label{fr-loss}
\end{equation}
where
\begin{equation}
	\mathcal{L}_\text{proxy-based} = \mathcal{L}_\text{pps} + \mathcal{L}_\text{pns} + \mathcal{L}_\text{pp}
	\label{proxy-based loss}
\end{equation}
Algorithm~\ref{algorithm2} demonstrate the training procedure for $\mathcal{L}_\text{proxy-based}$.

\subsection{Loss for Face Reconstruction}
\label{}
To improve the performance of FR, research was conducted on face reconstruction. The goal of this research is to enrich the model's feature representation in recognition tasks by utilizing auxiliary features, such as depth maps, during the reconstruction process. This approach aims to enhance FR accuracy and stability by leveraging three-dimensional information and albedo features.

In this study, a series of symbols and variables is used to describe the core process of face reconstruction. The input image $ \mathbf{I} \in \mathbb{R}^{3 \times W \times H} $ represents the three-channel input image, with width $ W $ and height $ H $. The canonical depth map $ \mathbf{d} \in \mathbb{R}^{W \times H} $ and the canonical albedo map $ \mathbf{a} \in \mathbb{R}^{3 \times W \times H} $ are generated through the modules $ \Phi_\mathbf{d} $ and $ \Phi_\mathbf{a} $, consisting of an encoder and a decoder.

The viewpoint matrix $ \omega \in \mathbb{R}^{2 \times 3} $ is used to describe the rotation and displacement of the camera's perspective. The lighting parameter vector $ \bm{l} = [k_a, k_d, l_{dx}, l_{dy}] \in \mathbb{R}^4 $ defines the ambient light intensity $ k_a $, the diffuse light intensity $ k_d $, and the light source direction $ \bm{l}_d = [l_{dx}, l_{dy}] $. Both the lighting parameter vector and the viewpoint matrix are generated through the modules $ \Phi_{\bm{l}} $ and $ \Phi_{\bm{v}} $. The lighting intensity map $ \mathbf{s} $ is calculated by using the formula $ \mathbf{s}_{x,y} = \max \{0, \langle \bm{l}_d, \mathbf{n}_{x,y} \rangle\} $, where $ \mathbf{n} $ is the normal vector calculated from the depth map. The normalized image $ \mathbf{I}_c $ is then computed from the albedo map and the lighting parameter map by using the formula $ \mathbf{I}_c = \mathbf{a} \circ (k_a + k_d \mathbf{s}) $.

During the reconstruction process, the camera intrinsic matrix $ \mathbf{K} $ is used along with the depth map's back-projection and transformation. The final reconstructed image is obtained through the transformation function $ \hat{\mathbf{I}} = \Pi(\mathbf{I}_c, \mathbf{d}, \omega, \mathbf{K}) $, where $ \bm{p}' $ and $ \bm{p} $ represent the transformed and original coordinates, respectively, satisfying $ \bm{p}' \propto \mathbf{K}(\mathbf{d} \cdot \mathbf{r} \mathbf{K}^{-1} \bm{p} + \bm{t}) $. The specific form of $ \mathbf{K} $ is a $3\times3$ matrix, containing camera parameters and the field of view angle $ \theta_{FOV} $. The expression is:
\begin{equation}
	\mathbf{K}=\left[\begin{array}{ccc}
		\frac{W-1}{2 \tan \frac{\theta_{F O V}}{2}} & 0 & \frac{W-1}{2} \\
		0 & \frac{W-1}{2 \tan \frac{\theta_{F O V}}{2}} & \frac{H-1}{2} \\
		0 & 0 & 1
	\end{array}\right]
\end{equation}

\begin{figure*}[h]
	\begin{center}
		\includegraphics[width=1.0\linewidth]{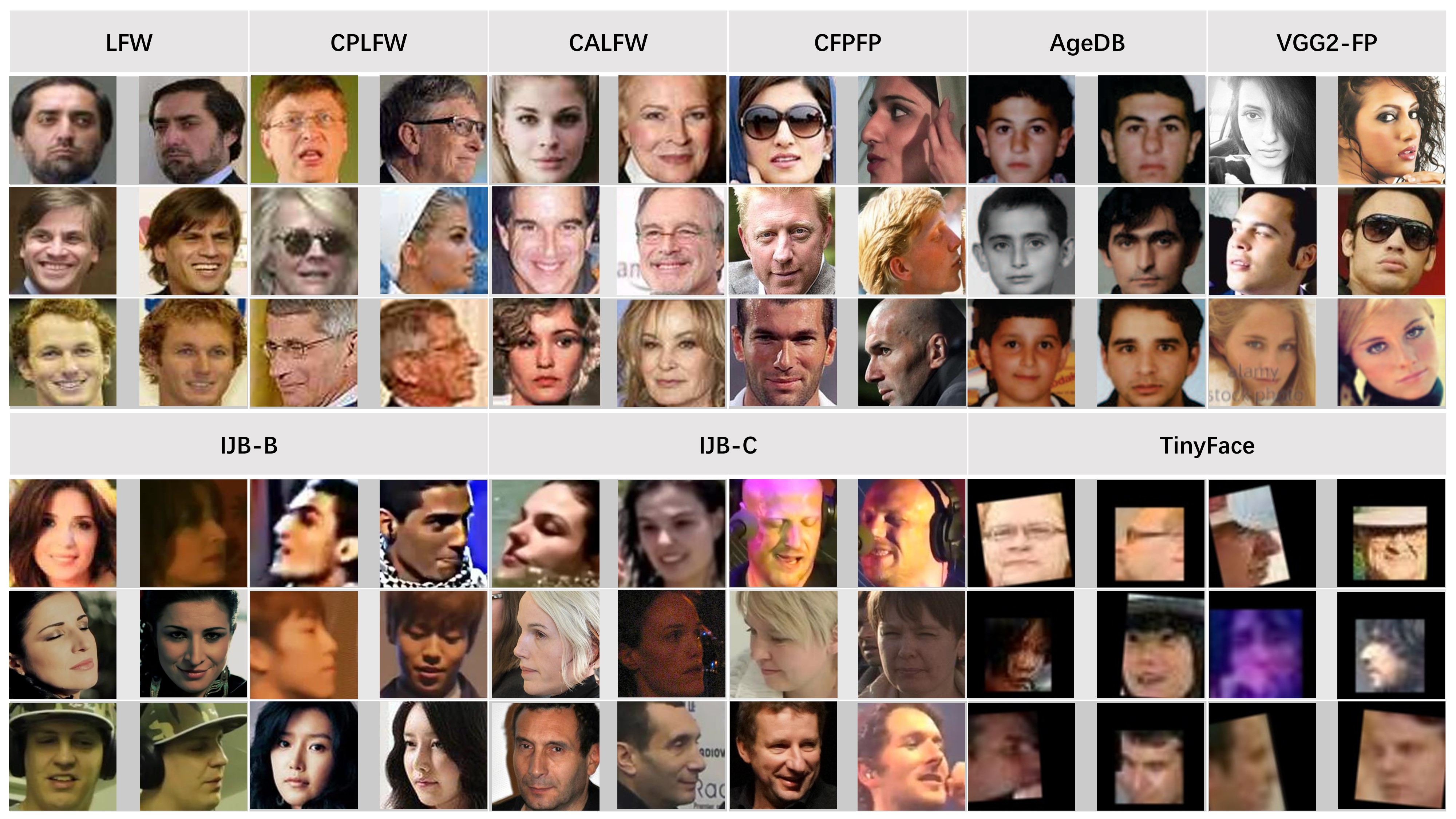}
	\end{center}
	\caption{The image shows a portion of the pictures included in the entire test set, revealing the style differences in the dataset images.}
	\label{datasets}
\end{figure*}

\begin{table*}[h]
	\caption{Ablation study with other methods. The red data in the table represent the best result, and the blue data represent the second place. For IJB-B and IJB-C, stringent verification standard was adopted, and the True Accept Rate (TAR) was reported at different False Accept Rates (FAR). TAR@FAR=$1e-6$ is reported. Closed-set rank retrieval (Rank-$20$) is used for TinyFace. RECO refers to the face reconstruction module. The parameter values we used are CosFace~\cite{wang2018cosface} ($m=0.35$), ArcFace~\cite{deng2019arcface} ($m=0.5$), MV-Arc-Softmax~\cite{wang2020mis} ($m=0.5$, $t=0.25$), SphereFaceRV2~\cite{spherefaceRV} ($m=1.4$), AdaFace~\cite{kim2022adaface} ($m=0.4$), and UCFace-ArcFace (Fine-tuned)~\cite{ucface} ($m=0.5$). LH$^2$Face represents UAMF + RECO + $\mathcal{L}_\text{proxy-based}$. All of the model data presented here are reproduced results.}
	\begin{center}
		\scriptsize
		(a) Performance comparison of recent methods on high-quality datasets with the ResNet18 backbone. \\
		\begin{tabular}{|l|c|c|c|cccccc|c|}
			\hline
			\multicolumn{1}{|c|}{\multirow{2}{*}{Method}} & \multicolumn{1}{c|}{\multirow{2}{*}{Venue}} & \multicolumn{1}{c|}{\multirow{2}{*}{Train Data}} & \multicolumn{1}{c|}{\multirow{2}{*}{Backbone}} & \multicolumn{7}{c|}{High-quality Datasets} \\ 
			\cline{5-11}
			\multicolumn{1}{|c|}{} & \multicolumn{1}{c|}{}  & \multicolumn{1}{c|}{}  & \multicolumn{1}{c|}{} & \multicolumn{1}{l}{\scalebox{0.85}{LFW~\cite {lfw}}} & \multicolumn{1}{l}{\scalebox{0.85}{CFP-FP~\cite {cfpfp}}} & \multicolumn{1}{l}{\scalebox{0.85}{CPLFW~\cite {cplfw}}} & \multicolumn{1}{l}{\scalebox{0.85}{AgeDB~\cite {agedb}}} & \multicolumn{1}{l}{\scalebox{0.85}{CALFW~\cite {calfw}}} & \multicolumn{1}{l|}{\scalebox{0.85}{VGG2-FP~\cite {vgg2fp}}} & \multicolumn{1}{c|}{\scalebox{0.85}{AVG}} \\ 
			\hline
			CosFace~\cite {wang2018cosface} & \multirow{2}{*}{\scalebox{0.85}{CVPR}} & \multirow{2}{*}{\scalebox{0.85}{subMS1MV2~\cite {deng2019arcface}}} & \multirow{2}{*}{\scalebox{0.85}{ResNet18~\cite{he2016deep}}} & $99.25$ & $89.06$ & $86.80$ & $95.10$ & $94.58$ & $90.20$ & $92.50$ \\
			CosFace + RECO  & & &  & $99.42$ & $90.31$ & $87.38$ & $95.42$ & $94.68$ & $91.42$ & $\bm{93.11}$ \\
			\hline
			
			ArcFace~\cite {deng2019arcface} & \multirow{2}{*}{\scalebox{0.85}{TPAMI}} & \multirow{2}{*}{\scalebox{0.85}{subMS1MV2~\cite {deng2019arcface}}} & \multirow{2}{*}{\scalebox{0.85}{ResNet18~\cite{he2016deep}}}  & $99.35$ & $89.94$ & $87.48$ &$95.50$ &$94.88$ &$91.14$ & $93.05$\\
			ArcFace + RECO & & & & $99.45$ & $90.17$ & $87.15$ &$95.38$ &$94.83$ &$91.58$ & $\bm{93.09}$\\
			\hline
			
			MV-Arc-Softmax~\cite {wang2020mis} & \multirow{2}{*}{\scalebox{0.85}{AAAI}} & \multirow{2}{*}{\scalebox{0.85}{subMS1MV2~\cite {deng2019arcface}}} & \multirow{2}{*}{\scalebox{0.85}{ResNet18~\cite{he2016deep}}} & $99.47$ & $88.47$ & $87.05$ & $94.20$ & $94.43$ & $90.22$ & $\bm{92.31}$ \\
			MV-Arc-Softmax + RECO & & & & $99.28$ & $88.07$ & $86.58$ & $94.85$ & $94.12$ & $90.60$ & $92.25$ \\
			\hline
			
			SphereFaceRV2~\cite {spherefaceRV} & \multirow{2}{*}{\scalebox{0.85}{TPAMI}} & \multirow{2}{*}{\scalebox{0.85}{subMS1MV2~\cite {deng2019arcface}}} & \multirow{2}{*}{\scalebox{0.85}{ResNet18~\cite{he2016deep}}} & $99.35$ & $93.03$ & $86.30$ & $92.97$ & $94.10$ & $92.54$ & $93.05$ \\
			SphereFaceRV2 + RECO & & & & $99.48$ & $93.27$ & $87.12$ & $93.72$ & $94.20$ & $92.84$ & $\bm{93.44}$ \\
			\hline
			
			AdaFace~\cite {kim2022adaface} & \multirow{2}{*}{\scalebox{0.85}{CVPR}} & \multirow{2}{*}{\scalebox{0.85}{subMS1MV2~\cite {deng2019arcface}}} & \multirow{2}{*}{\scalebox{0.85}{ResNet18~\cite{he2016deep}}} & $99.33$ & $88.79$ & $87.13$ & $95.18$ & $94.42$ & $91.04$ & $92.65$ \\
			AdaFace + RECO & & & & $99.43$ & $89.01$ & $87.37$ & $95.25$ & $94.78$ & $90.52$ & $\bm{92.73}$ \\
			\hline
			
			UCFace-ArcFace~\cite {ucface} & \multirow{2}{*}{\scalebox{0.85}{TIFS}} & \multirow{2}{*}{\scalebox{0.85}{subMS1MV2~\cite {deng2019arcface}}} & \multirow{2}{*}{\scalebox{0.85}{ResNet18~\cite{he2016deep}}} & $99.30$ & $90.37$ & $87.28$ & $95.13$ & $94.92$ & $91.16$ & $93.03$ \\
			UCFace-ArcFace + RECO & & & & $99.35$ & $90.94$ & $87.43$ & $94.83$ & $94.67$ & $91.94$ & $\bm{93.19}$ \\
			\hline \hline
			
			UAMF & & \multirow{4}{*}{\scalebox{0.85}{subMS1MV2~\cite {deng2019arcface}}} & \multirow{4}{*}{\scalebox{0.85}{ResNet18~\cite{he2016deep}}} & $99.48$ & $91.61$ & $88.10$ & $94.98$ & $94.48$ & $92.48$ & $93.52$ \\
			UAMF + RECO & & &  & $99.38$ & $91.93$ & $87.98$ & $94.88$ & $94.63$ & $92.42$ & $93.54$ \\
			UAMF + $\mathcal{L}_\text{proxy-based}$ & & & & $99.53$ & $92.70$ & $88.03$ & $94.95$ & $94.37$ & $92.22$ & $\second{93.63}$ \\
			LH$^2$Face & & & & $99.47$ & $92.27$ & $87.85$ & $94.75$ & $94.80$ & $93.16$ & $\first{93.72}$ \\
			\hline
		\end{tabular}
		\scriptsize
	\end{center}
	\begin{center}
		\scriptsize
		(b) Performance comparison of recent methods on IJB-B, IJB-C, and TinyFace datasets with the ResNet18 backbone. \\
		\begin{tabular}{|l|c|c|c|cc|c|c|}
			\hline
			\multicolumn{1}{|c|}{\multirow{2}{*}{Method}} & \multicolumn{1}{c|}{\multirow{2}{*}{Venue}} & \multicolumn{1}{c|}{\multirow{2}{*}{Train Data}} & \multicolumn{1}{c|}{\multirow{2}{*}{Backbone}} & \multicolumn{1}{c}{IJB-B~\cite {ijbb}} & \multicolumn{1}{c}{IJB-C~\cite {ijbc}} &  & \multicolumn{1}{c|}{\multirow{2}{*}{TinyFace~\cite {tinyface}}} \\
			\cline{5-7}
			& & & &  \multicolumn{1}{c}{$1e-6$} & \multicolumn{1}{c|}{$1e-6$} & \multicolumn{1}{c|}{AVG}  & \\ 
			\hline
			CosFace~\cite {wang2018cosface}& \multirow{2}{*}{CVPR} & \multirow{2}{*}{subMS1MV2~\cite {deng2019arcface}} & \multirow{2}{*}{ResNet18~\cite{he2016deep}} & $34.36$ & $76.28$ & $55.32$ & $\second{68.80}$ \\
			CosFace + RECO & & & & $34.97$ & $78.35$ & $\bm{56.66}$ & $67.92$ \\
			\hline
			
			ArcFace~\cite {deng2019arcface} & \multirow{2}{*}{TPAMI} & \multirow{2}{*}{subMS1MV2~\cite {deng2019arcface}} & \multirow{2}{*}{ResNet18~\cite{he2016deep}} & $35.13$ & $78.92$ & $57.03$ & $\bm{68.05}$ \\
			ArcFace + RECO & & & & $39.37$ & $79.37$ & $\second{59.37}$ & $67.38$ \\
			\hline
			
			MV-Arc-Softmax~\cite {wang2020mis} & \multirow{2}{*}{AAAI} & \multirow{2}{*}{subMS1MV2~\cite {deng2019arcface}} & \multirow{2}{*}{ResNet18~\cite{he2016deep}} & $32.06$ & $77.73$ & $54.90$ & $\first{69.31}$ \\
			MV-Arc-Softmax + RECO & & & & $34.95$ & $77.11$ & $\bm{56.03}$ & $68.16$ \\
			\hline
			
			SphereFaceRV2~\cite {spherefaceRV} & \multirow{2}{*}{TPAMI} & \multirow{2}{*}{subMS1MV2~\cite {deng2019arcface}} & \multirow{2}{*}{ResNet18~\cite{he2016deep}} & $30.75$ & $62.32$ & $\bm{46.54}$ & $59.17$ \\
			SphereFaceRV2 + RECO & & & & $31.23$ & $60.36$ & $45.80$ & $\bm{59.44}$ \\
			\hline
			
			AdaFace~\cite {kim2022adaface} & \multirow{2}{*}{CVPR} & \multirow{2}{*}{subMS1MV2~\cite {deng2019arcface}} & \multirow{2}{*}{ResNet18~\cite{he2016deep}} &  $33.66$ & $77.98$ & $55.82$ & $\bm{68.64}$ \\
			AdaFace + RECO & & & &  $35.99$ & $79.12$ & $\bm{57.56}$ & $68.24$\\
			\hline
			
			UCFace-ArcFace~\cite {ucface} & \multirow{2}{*}{TIFS} & \multirow{2}{*}{subMS1MV2~\cite {deng2019arcface}} & \multirow{2}{*}{ResNet18~\cite{he2016deep}} &  $38.64$ & $74.16$ & $56.40$ & $67.09$\\
			UCFace-ArcFace + RECO & & & &  $40.53$ & $74.50$ & $\bm{57.52}$ & $\bm{67.41}$\\
			\hline \hline
			
			UAMF & & \multirow{4}{*}{subMS1MV2~\cite {deng2019arcface}} & \multirow{4}{*}{ResNet18~\cite{he2016deep}} &  $40.78$ & $76.05$ & $58.42$ & $66.28$ \\
			UAMF + RECO & & & &  $39.38$ & $78.00$ & $58.69$ & $65.77$ \\
			UAMF + $\mathcal{L}_\text{proxy-based}$ & & & &  $43.26$ & $74.54$ & $58.90$ & $66.79$ \\
			LH$^2$Face & & & &  $43.53$ & $75.27$ & $\first{59.40}$ & $\bm{67.46}$ \\
			\hline
		\end{tabular}
		\scriptsize
	\end{center}
	\label{table:SoTA}
\end{table*}

\begin{table*}[h]
	\caption{Comparison with recent state-of-the-art(SoTA) methods. All parameter values remain the same as in the previous ablation experiments. Apart from LH$^2$Face, all other model data are derived from the original paper's data or model files, meaning they are not reproduced results.}
	\begin{center}
		\scriptsize
		(a) Performance comparison of recent methods on high-quality datasets with the ResNet101 backbone. \\
		\begin{tabular}{|l|c|c|c|cccccc|c|}
			\hline
			\multicolumn{1}{|c|}{\multirow{2}{*}{Method}} & \multicolumn{1}{c|}{\multirow{2}{*}{Venue}} & \multicolumn{1}{c|}{\multirow{2}{*}{Train Data}} & \multicolumn{1}{c|}{\multirow{2}{*}{Backbone}} & \multicolumn{7}{c|}{High-quality Datasets} \\ 
			\cline{5-11}
			\multicolumn{1}{|c|}{} & \multicolumn{1}{c|}{}  & & & \multicolumn{1}{l}{\scalebox{0.85}{LFW~\cite {lfw}}} & \multicolumn{1}{l}{\scalebox{0.85}{CFP-FP~\cite {cfpfp}}} & \multicolumn{1}{l}{\scalebox{0.85}{CPLFW~\cite {cplfw}}} & \multicolumn{1}{l}{\scalebox{0.85}{AgeDB~\cite {agedb}}} & \multicolumn{1}{l}{\scalebox{0.85}{CALFW~\cite {calfw}}} & \multicolumn{1}{l|}{\scalebox{0.85}{VGG2-FP~\cite {vgg2fp}}} & \multicolumn{1}{c|}{\scalebox{0.85}{AVG}} \\ 
			\hline
			
			CosFace~\cite {wang2018cosface} & CVPR18 & MS1MV2~\cite {deng2019arcface} & ResNet101~\cite{he2016deep} & $99.81$ & $98.12$ & $92.28$ & $98.11$ & $95.76$ & - & - \\
			
			MV-Arc-Softmax~\cite {wang2020mis} & AAAI20 & MS1MV2~\cite {deng2019arcface} & ResNet101~\cite{he2016deep} & $99.80$ & $98.28$ & $92.83$ & $97.95$ & $96.10$ & - & - \\
			
			CurricularFace~\cite {huang2020curricularface} & CVPR20 & MS1MV2~\cite {deng2019arcface} & ResNet101~\cite{he2016deep} & $99.80$ & $98.37$ & $93.13$ & $98.32$ & $\bm{96.20}$ & - & - \\
			
			SCF-ArcFace~\cite {li2021spherical} & CVPR21 & MS1MV2~\cite {deng2019arcface} & ResNet101~\cite{he2016deep} & $99.82$ & $98.40$ & $93.16$ & $98.30$ & $96.12$ & - & - \\
			
			MagFace~\cite {meng2021magface} & CVPR21 & MS1MV2~\cite {deng2019arcface} & ResNet101~\cite{he2016deep} & $\bm{99.83}$ & $98.46$ & $92.87$ & $98.17$ & $96.15$ & - & - \\
			
			ArcFace~\cite {deng2019arcface} & TPAMI22 & MS1MV2~\cite {deng2019arcface} & ResNet101~\cite{he2016deep}  & $99.83$ & $98.27$ & $92.08$ &$98.28$ &$95.45$ & - & - \\
			
			MvCoM-CosFace~\cite {liu2022diverse_data} & CVPR22 & MS1MV2~\cite {deng2019arcface} & ResNet101~\cite{he2016deep} & $99.80$ & $98.37$ & $92.75$ & - & - & - & -\\
			
			AdaFace~\cite {kim2022adaface} & CVPR22 & MS1MV2~\cite {deng2019arcface} & ResNet101~\cite{he2016deep} & $99.82$ & $98.49$ & $93.53$ & $98.05$ & $96.08$ & $95.70$ & $96.95$ \\
			
			DDC~\cite{ddc} & TPAMI23 & MS1MV2~\cite {deng2019arcface} & ResNet101~\cite{he2016deep} & $99.80$ & $98.30$ & $92.30$ & $98.00$ & $96.00$ & - & - \\ 
			
			FaceT-B~\cite{facet-b} & TIFS23 & MS1MV2~\cite {deng2019arcface} & ResNet101~\cite{he2016deep} & $99.82$ & $98.23$ & $92.62$ & $98.18$ & $95.68$ & - & - \\ 

			QSD-ArcFace~\cite {qsd} & TMM24 & MS1MV2~\cite {deng2019arcface} & ResNet101~\cite{he2016deep} & $99.80$ & $98.61$ & $93.16$ & - & - & - & -\\
			
			FRABSM~\cite {FRABSM} & KBS24 & MS1MV2~\cite {deng2019arcface} & ResNet101~\cite{he2016deep} & $99.82$ & $98.56$ & $93.45$ & $98.32$ & $96.12$ & $95.48$ & $\second{96.96}$ \\ 
			
			CoReFace~\cite{coreface} & PR24 & MS1MV2~\cite {deng2019arcface} & ResNet101~\cite{he2016deep} & $\bm{99.83}$ & $98.60$ & $93.27$ & $\bm{98.37}$ & $\bm{96.20}$ & - & - \\ 
			
			IIC-AdaFace~\cite{iic} & ICLR24 & MS1MV2~\cite {deng2019arcface} & ResNet101~\cite{he2016deep} & $99.78$ & $98.41$ & $93.48$ & $98.12$ & $96.18$ & $95.80$ & $\second{96.96}$ \\ 
			
			\hdashline
			
			LH$^2$Face & & MS1MV2~\cite {deng2019arcface} & ResNet101~\cite{he2016deep} & $99.82$ & $\bm{99.03}$ & $\bm{94.07}$ & $97.30$ & $95.88$ & $\bm{96.46}$ & $\first{97.09}$ \\
			\hline
		\end{tabular}
		\scriptsize
	\end{center}
	\begin{center}
		\scriptsize
		(b) Performance comparison of recent methods on IJB-B, IJB-C, and TinyFace datasets with the ResNet101 backbone. \\
		\begin{tabular}{|l|c|c|c|cc|c|c|}
			\hline
			\multicolumn{1}{|c|}{\multirow{2}{*}{Method}} & \multicolumn{1}{c|}{\multirow{2}{*}{Venue}} & \multicolumn{1}{c|}{\multirow{2}{*}{Train Data}} & \multicolumn{1}{c|}{\multirow{2}{*}{Backbone}} & \multicolumn{1}{c}{IJB-B~\cite {ijbb}} & \multicolumn{1}{c}{IJB-C~\cite {ijbc}} &  & \multicolumn{1}{c|}{\multirow{2}{*}{TinyFace~\cite {tinyface}}} \\
			\cline{5-7}
			& & & &  \multicolumn{1}{c}{$1e-6$} & \multicolumn{1}{c|}{$1e-6$} & \multicolumn{1}{c|}{AVG}  & \\ 
			\hline
			CosFace~\cite {wang2018cosface} & CVPR18 & MS1MV2~\cite {deng2019arcface} & ResNet101~\cite{he2016deep} & $40.41$ & $87.96$ & $64.19$ & - \\
			
			MagFace~\cite {meng2021magface} & CVPR21 & MS1MV2~\cite {deng2019arcface} & ResNet101~\cite{he2016deep} & $42.32$ & $90.24$ & $66.28$ & - \\
			
			ArcFace~\cite {deng2019arcface} & TPAMI22 & MS1MV2~\cite {deng2019arcface} & ResNet101~\cite{he2016deep} & $38.68$ & $85.65$ & $62.17$ & - \\
			
			3D-BERL~\cite {3d-berl} & CVPR22 & MS1MV2~\cite {deng2019arcface} & ResNet101~\cite{he2016deep} & $45.77$ & $88.45$ & $67.11$ & - \\
			
			AdaFace~\cite {kim2022adaface} & CVPR22 & MS1MV2~\cite {deng2019arcface} & ResNet101~\cite{he2016deep} &  $46.78$ & $89.74$ & $\second{68.26}$ & $\second{74.17}$ \\
			
			FRABSM~\cite {FRABSM} & KBS24 & MS1MV2~\cite {deng2019arcface} & ResNet101~\cite{he2016deep} &  $46.28$ & $86.52$ & $66.40$ & $73.47$ \\
			
			CoReFace~\cite{coreface} & PR24 & MS1MV2~\cite {deng2019arcface} & ResNet101~\cite{he2016deep} &  $47.02$ & $89.34$ & $68.18$ & - \\
			
			IIC-AdaFace~\cite{iic} & ICLR24 & MS1MV2~\cite {deng2019arcface} & ResNet101~\cite{he2016deep} &  - & $89.99$ & - & - \\
			
			Joint-BERL~\cite{joint-berl} & TIP24 & MS1MV2~\cite {deng2019arcface} & ResNet101~\cite{he2016deep} &  $45.80$ & $89.42$ & $67.61$ & - \\
			
			\hdashline
			
			LH$^2$Face & & MS1MV2~\cite {deng2019arcface} & ResNet101~\cite{he2016deep} &  $49.39$ & $88.01$ & $\first{68.70}$ & $\first{74.68}$ \\
			\hline
		\end{tabular}
		\scriptsize
	\end{center}
	\label{table:SoTA2}
\end{table*}

In the field of FR, 3D-BERL~\cite{3d-berl} proposed a method based on face reconstruction. The paper extracts the viewpoint matrix and illumination parameter vector from the original image, and combines them with the canonical depth map and canonical albedo map obtained from the feature map of the third block to gradually reconstruct the face. The L1 distance between the reconstructed face and the original image is incorporated into the loss function to optimize the reconstruction performance.

However, although the method claims that its 'subsequent layer (Stage 4) can overcome the influence of pose and lighting to obtain more robust face embeddings', this argument appears to be insufficiently reliable. Therefore, we aim to utilize the information provided by face reconstruction more effectively, not just as an auxiliary bypass.
The 3D-BERL framework only indirectly uses 3D information to influence the training of the recognition network, failing to integrate it effectively into the final face embedding. In contrast, the 3D features—i.e., the depth map and canonical face image—are directly integrated into the input of the FR task, directly improving recognition performance.

In the face reconstruction process, our loss function combines several loss terms, with the FR loss being the dominant term, while other auxiliary losses play a balanced and supportive role. The specific loss function is as follows:
\begin{equation}
	\mathcal{L}_\text{train} = \mathcal{L}_\text{FR} + \lambda_\text{reco} \mathcal{L}_\text{reco} + \lambda_\text{canon} \mathcal{L}_\text{FR}^\text{canon} + \lambda_\text{view} \mathcal{L}_\text{view}
\end{equation}
In this loss function, $\mathcal{L}_\text{FR}$ is the main loss term, responsible for optimizing FR performance.
The value of $\lambda_\text{reco}$ is set to $0.01$ because the reconstruction network structure is more complex and requires a lower weight for convergence.
The values of $\lambda_\text{canon}$ and $\lambda_\text{view}$ are both $0.001$, serving as auxiliary losses to improve multi-view reconstruction and stability through canonical image feature matching and multi-view variance constraints, respectively.
The specific expression of the loss function is provided in the appendix.

\section{Experiment and Result}
\label{sec:experiment}

\subsection{Datasets and Implementation Details}

In our study, the CelebA~\cite{celeba} training set was used for face reconstruction, and a subset of MS1MV2~\cite{deng2019arcface}, referred to as subMS1MV2, was used for FR training. The subMS1MV2 subset contains approximately $1/8$ of the images in MS1MV2. These datasets provided us with a rich set of training samples.
The performance was evaluated on several different test sets to validate the effectiveness of the method. Among these datasets, LFW~\cite{lfw}, CFP-FP~\cite{cfpfp}, CPLFW~\cite{cplfw}, AgeDB~\cite{agedb}, CALFW~\cite{calfw}, and VGG2-FP~\cite{vgg2fp} are the most widely used. The image quality of these datasets is high, including both hard and easy images, enabling accurate evaluation of the model's performance.
The IJB-B and IJB-C~\cite {ijbb, ijbc} contain a small number of low-quality images and a large number of hard high-quality images.
TinyFace is a low-quality dataset without high-quality face images.
The images included in these test sets can be found in Fig.~\ref{datasets}.

CelebA is a large-scale real-world face dataset containing over 200,000 real face images with bounding boxes. We use MTCNN to perform coarse cropping around the head region and employ the official training/validation/testing splits. For face reconstruction, the Adam optimizer is used to train batches of input images, each with a size of $64 \times 64$ pixels, for a total of $30$ epochs.

When it comes to FR training, we adopted and followed the procedure used in ArcFace~\cite {deng2019arcface}. First, fine cropping was performed on the face images operated by MTCNN, and they were aligned based on five preset landmarks. This resulted in unified face images of size $112 \times 112$. For the main neural network architecture, i.e., backbone network, the improved version of ResNet~\cite{he2016deep} provided by ArcFace was referred to and used. The model was trained for $20$ epochs with the MultiStepLr optimizer. A dynamic adjustment strategy was employed to determine the learning rate. In UAMF, $ n $ is set to $256$, $ d $ is $512$, and $ \tau $ is set to $1.0$. In the reproduced UCFace~\cite{ucface}, $ d_f = 512 $, $ d_g = 256 $, and $ \tau = 1.0 $ were used. Additionally, proxy-sample was used rather than sample-sample as the source of similarity for $ \mathcal{L}_U $. The initial learning rate was set to $0.1$, and during training, the learning rate was halved every two epochs for better training results.
For data augmentation, three commonly used techniques in image classification tasks were implemented: cropping, resizing, and geometric occlusion. To improve the model's generalization ability, these techniques were applied with a probability of $0.2$. Finally, the batch size for training was set to $128$. In terms of hardware, an NVIDIA GeForce RTX 4090 server was used to train the ResNet18 network, and an NVIDIA A100-SXM4-40GB was used to train ResNet101. All models were trained with 16-bit precision.

The performance metrics include the average $1\colon1$ verification accuracy on LFW, CFP-FP, CPLFW, AgeDB, CALFW, and VGG2-FP. The $1\colon1$ verification accuracy is an important metric for evaluating the performance of FR models. It refers to the proportion of correct judgments made by the model on whether two images belong to the same person in a one-to-one verification task. In these evaluations, each image pair is labeled as either "match" or "non-match," and the model must determine whether the pair's relationship is correctly identified. By calculating the average $1\colon1$ verification accuracy across the LFW, CFP-FP, CPLFW, AgeDB, CALFW, and VGG2-FP datasets, the model's overall performance in handling high-quality images under different environments and conditions can be assessed.

For the IJB-B and IJB-C evaluations, stringent verification standard was adopted, and the TAR at different FAR, specifically TAR@FAR=$1e-6$, was reported. TAR@FAR is an important metric for assessing the performance of FR systems, especially in applications with high security requirements. These metrics help us understand the system's performance at very low false acceptance rates, ensuring the system's security and reliability.

For the TinyFace dataset, the closed-set ranking retrieval method was used to evaluate the model's recognition ability, specifically Rank-$20$. Rank-$20$ refers to the accuracy of the target image being recognized as one of the top twenty matches, providing an assessment of the model's recognition capability. The closed-set ranking retrieval method evaluates the model's accuracy in recognizing low-quality face images by ranking the model's performance within a closed candidate set.

The TAR@FAR evaluation on the IJB-B and IJB-C datasets, as well as the closed-set ranking retrieval evaluation on the TinyFace dataset, provide multi-faceted performance metrics that help to comprehensively understand and compare the strengths and weaknesses of different FR algorithms. We will assess the algorithm's performance on hard high-quality datasets, including LFW, CFP-FP, CPLFW, AgeDB, CALFW, VGG2-FP, IJB-B, and IJB-C. The goal is to improve accuracy on these test sets, which is also the main objective in this paper. The performance on TinyFace only needs to avoid a significant decline.

\begin{figure*}[h]
	\begin{center}
		\includegraphics[width=1.0\linewidth]{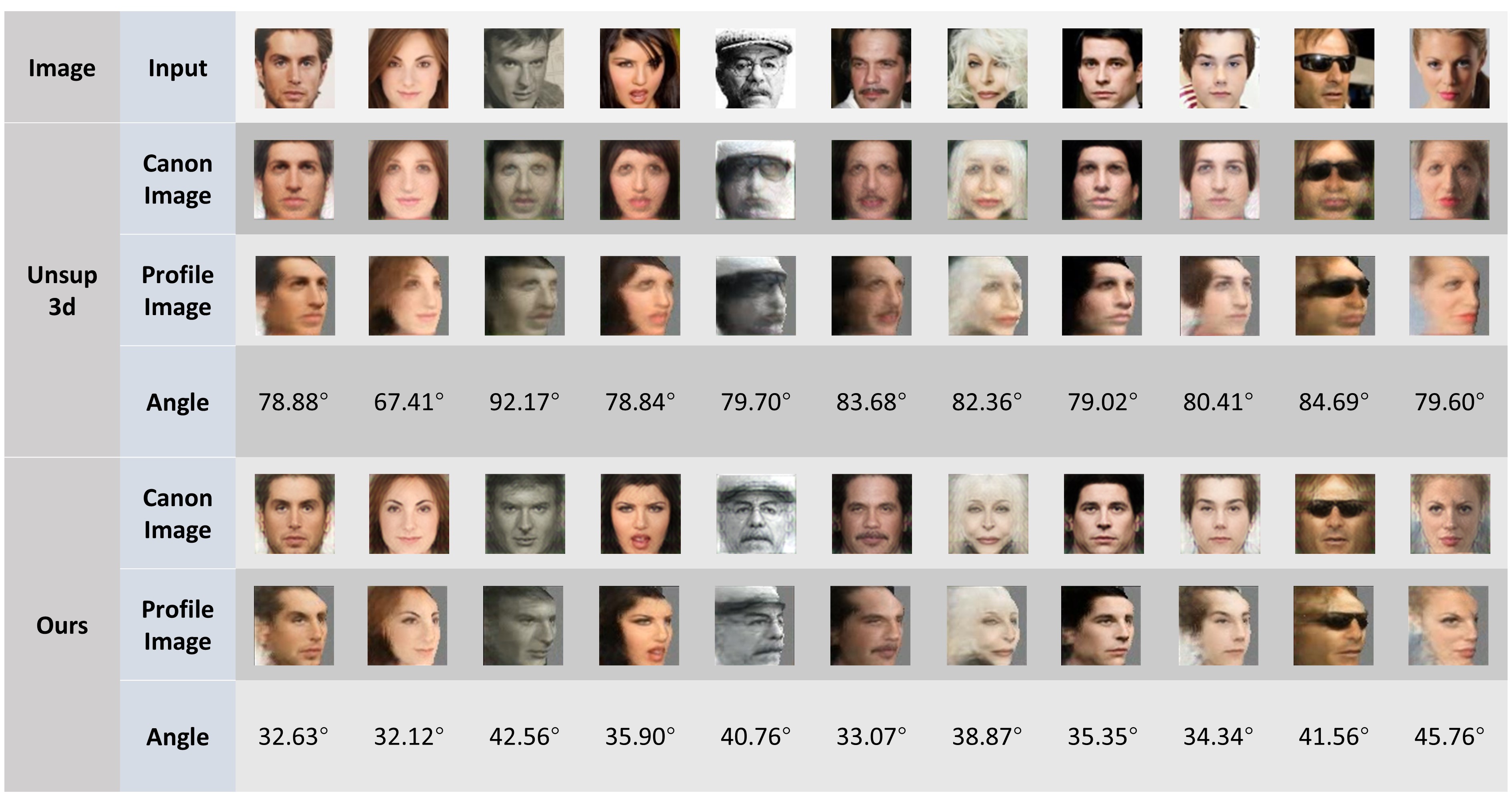}
	\end{center}
	\caption{This picture compares our method with Unsup3d~\cite{unsup3d}. The "Profile image" refers to the image obtained by rotating the canonical image by a fixed angle. "Angle" represents the angle between the feature vectors of the input face image and the canonical face image, measured in degrees.}
	\label{picture}
\end{figure*}

\begin{table}
	\caption{Comparison of ablation experiments on elements $\mathcal{L}_\text{pps}$, $\mathcal{L}_\text{pp}$, and $\mathcal{L}_\text{pns}$ in the ArcFace.}
	\begin{center}
		\scriptsize
		\begin{tabular}{|l|c|c|c|ccc|}
			\hline
			\multicolumn{1}{|c|}{Method} & \multicolumn{1}{c|}{$\mathcal{L}_\text{pps}$} & \multicolumn{1}{c|}{$\mathcal{L}_\text{pp}$} & \multicolumn{1}{c|}{$\mathcal{L}_\text{pns}$} & High-quality & IJB-B/C & TinyFace \\ 
			\hline
			
			& $w/o$ & $w/o$ & $w/o$ & $93.05$ & $57.03$ &$\second{68.05}$\\
			ArcFace~\cite {deng2019arcface} & $w/$ & $w/o$ & $w/o$ & $\first{93.19}$ & $58.55$ &$67.62$\\
			($m=0.5$)& $w/$ & $w/$ & $w/o$ & $93.09$ & $\second{58.62}$ &$\first{68.48}$\\
			& $w/$ & $w/$ & $w/$ & $\second{93.16}$ & $\first{59.39}$ &$67.81$\\
			\hline
		\end{tabular}
		\\
		\scriptsize
	\end{center}
	\label{table:ablation}
\end{table}

\subsection{The Performance of UAMF}
As shown in Table~\ref{table:SoTA}, LH$^2$Face with only UAMF outperforms other methods on high-quality datasets, and also shows excellent performance on IJB-B and IJB-C. Among them, CPLFW achieves the best performance, surpassing all other methods. This could be because the hardness of CPLFW is the highest among the six high-quality datasets. The principle of UAMF helps us focus on those hard high-quality samples during training, matching the characteristics of the CPLFW, IJB-B, and IJB-C datasets.

In the TinyFace dataset, our performance is somewhat less impressive compared to other datasets. This is largely due to the lower quality of these images, which results in smaller feature norms that are less emphasized during training. Since our primary focus is not on low-quality images, where the available information is inherently limited, achieving exceptionally high accuracy on them was not our main objective. Even using deblurring methods, it cannot guarantee that the restored result will be correct. There may be more than one optimal solution.

\subsection{The Performance of Proxy-based Loss}
An ablation study on the proxy-based loss was conducted, shown in Table~\ref{table:ablation}. It is clear that this loss improves performance on high-quality and IJB-B, IJB-C, but has no positive effect on TinyFace. Certainly, since our goal is not to recognize low-quality face images but to identify hard high-quality images, this performance drop is acceptable. In the last row of Table~\ref{table:SoTA} (a) and (b), $\mathcal{L}_\text{proxy-based}$ was added to the training, resulting in a certain performance improvement. Here, $\lambda_\text{pps} = 5$, $\lambda_\text{pp} = 150$, and $\lambda_\text{pns} = 20$. Although these values for $\lambda_\text{pps}$, $\lambda_\text{pp}$, and $\lambda_\text{pns}$ seem large, when trained until the final epoch, the values of $\mathcal{L}_\text{pps}$, $\mathcal{L}_\text{pp}$, and $\mathcal{L}_\text{pns}$ were all less than $0.2$, while the core loss $\mathcal{L}_\text{vMF}$ remained relatively large. Therefore, the auxiliary losses do not produce any significant interference to the main loss or lead to overfitting.

\subsection{The Impact of Incorporating Face Reconstruction on Face Recognition}
In this experiment, the results of face reconstruction were incorporated as part of the input by modifying the input channels of the original FR network. Specifically, the number of input channels of the Input\_layer network was increased from $3$ to $7$. The original network only accepts three-channel images as input, but now the canonical face image and depth map from the face reconstruction process are additionally input. The canonical face image is a three-channel image, while the depth map is a single-channel image. Apart from modifying the input layer, the rest of the network architecture remains unchanged.

After training the FR network, a noticeable improvement in recognition accuracy on the IJB-B and IJB-C datasets was observed. These datasets contain hard face images from various poses, and incorporating face reconstruction information effectively enhances recognition performance. There was also a slight improvement in the high-quality datasets, possibly due to more obvious performance bottlenecks in those datasets.

On the TinyFace dataset, the face reconstruction method did not show significant improvements as expected. The reason lies in the fact that face reconstruction performs poorly on low-quality images. Specifically, in some images from these datasets, due to extreme poses and poor quality, the reconstruction results could not provide useful information, negatively impacted FR performance. This reflects the limitations of face reconstruction when dealing with low-quality images.

Overall, the performance of this method on multi-pose datasets demonstrates the potential of incorporating face reconstruction as part of the input in improving FR accuracy, especially by combining canonical face and depth information to provide more valuable cues.

\subsection{The Impact of Incorporating Face Recognition on Face Reconstruction}

\begin{figure}[h]
	\begin{center}
		\includegraphics[width=1.0\linewidth]{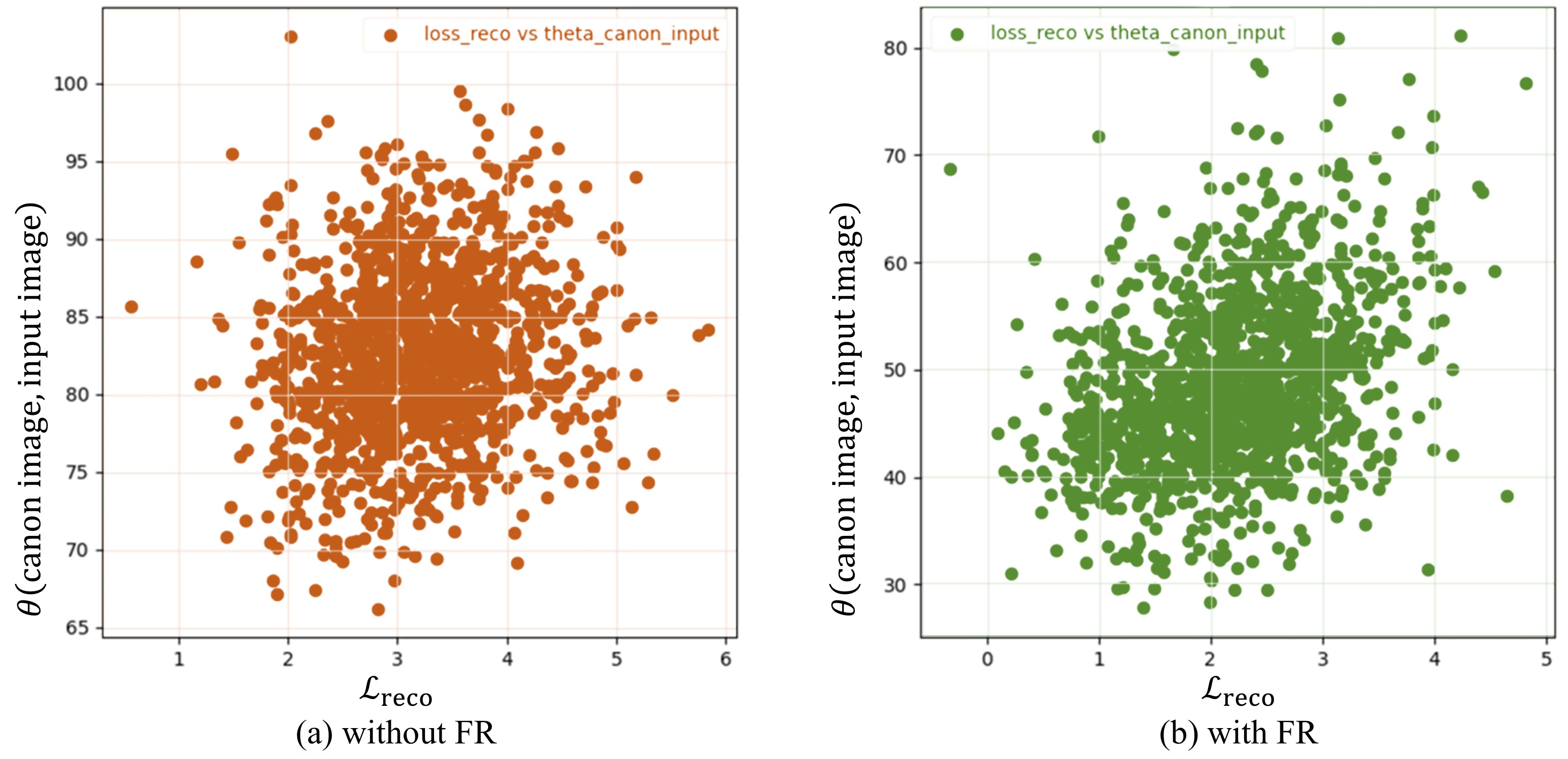}
	\end{center}
	\caption{The vertical axis represents the angle between the feature vectors of the input face image and the reconstructed canon face image. The horizontal axis represents the reconstruction loss value of the input face image. It can be found that (a) shows the relationship graph before incorporating FR, while (b) shows the relationship graph after incorporating FR.}
	\label{input_canon}
\end{figure}

After introducing FR into the face reconstruction task, it was observed that the reconstructed face images are more likely to be recognized as original images belonging to the same person. This similarity is primarily manifested at the semantic level, rather than merely in terms of shape similarity. Specifically, FR is not just a simple shape matching process, but rather a higher-level perceptual loss. Therefore, FR can be viewed as an advanced perceptual guide, helping improve face reconstruction quality.

Fig.~\ref{picture} shows the angle of similarity between the reconstructed canonical face image and the original image in terms of FR. To facilitate the understanding, we converted the angle from radians to degrees. The results indicate that after auxiliary training with FR, the angle of similarity between the reconstructed face and the original image significantly decreases. This suggests that face reconstruction not only approximates the shape of the original image, but also maintains a high level of semantic similarity.

In addition, Fig.~\ref{input_canon} further validates this. The former shows the angle of similarity before incorporating FR, while the latter says the changes after its inclusion. Compared to Fig.~\ref{input_canon} (a), the angle in Fig.~\ref{input_canon} (b) decreases significantly, indicating that the similarity between the reconstructed face image and the original image in the FR task is notably improved after the introduction of FR.

More importantly, Fig.~\ref{input_canon} (b) reveals a key phenomenon: after incorporating FR, a weak correlation appears between the reconstruction loss and the angle between feature vectors. Specifically, images with larger reconstruction losses tend to have larger angles, which suggests a certain degree of correlation between the increase in reconstruction loss and the decrease in reconstruction quality. Before the FR training was introduced, this correlation did not exist, further supporting the positive guiding role of FR in the reconstruction process.

In summary, introducing FR into the face reconstruction task not only effectively improves the reconstruction quality, enhancing the semantic similarity between the reconstructed face and the original image, but also establishes a correlation between the reconstruction loss and the FR similarity. This demonstrates the importance and effectiveness of FR in this task.

\subsection{Comparison with SoTA Methods}

To compare with SoTA methods, we trained LH$^2$Face based on ResNet101 and the complete MS1MV2 dataset. Due to limited GPU memory during training, we used a batch size of 256 instead of 512, and, as in the ablation experiments, trained for only 20 epochs. The detailed data can be found in Table~\ref{table:SoTA2}. It can be observed that LH$^2$Face performs well on datasets involving large poses, such as CFP-FP, CPLFW, and VGG2-FP, but is slightly inferior on cross-age testing sets. At the same time, LH$^2$Face also shows good performance on IJB-B and IJB-C.

\section{Conclusion}
In this paper, a method based on the margin and the vMF distribution was proposed. By defining an adaptive concentration, the performance of FR algorithms in recognizing hard high-quality samples was enhanced. We performed AD visualizations and ablation experiments to confirm the high effectiveness of the designed proxy-based loss. Additionally, face reconstruction was also incorporated as an auxiliary method to improve FR. The approach presented offered a new perspective, and the experimental results effectively support our research.

There are still challenges in handling low quality images, particularly addressing issues such as deblurring. Looking ahead, alternative reconstruction methods will be explored, such as diffusion models or GANs, which provides valuable insights into enhancing both the reconstruction and recognition processes.

%
%
%

\bibliographystyle{IEEEtran}
\bibliography{reference}

@inproceedings{wang2017normface,
	title={Normface: L2 hypersphere embedding for face verification},
	author={Wang, Feng and Xiang, Xiang and Cheng, Jian and Yuille, Alan Loddon},
	booktitle={Proceedings of the 25th ACM international conference on Multimedia},
	pages={1041--1049},
	year={2017}
}

@INPROCEEDINGS{liu2017sphereface,
	author={Liu, Weiyang and Wen, Yandong and Yu, Zhiding and Li, Ming and Raj, Bhiksha and Song, Le},
	booktitle={2017 IEEE Conference on Computer Vision and Pattern Recognition (CVPR)}, 
	title={SphereFace: Deep Hypersphere Embedding for Face Recognition}, 
	year={2017},
	volume={},
	number={},
	pages={6738-6746}}

@INPROCEEDINGS{wang2018cosface,
	author={Wang, Hao and Wang, Yitong and Zhou, Zheng and Ji, Xing and Gong, Dihong and Zhou, Jingchao and Li, Zhifeng and Liu, Wei},
	booktitle={2018 IEEE/CVF Conference on Computer Vision and Pattern Recognition}, 
	title={CosFace: Large Margin Cosine Loss for Deep Face Recognition}, 
	year={2018},
	volume={},
	number={},
	pages={5265-5274}}

@ARTICLE{deng2019arcface,
	author={Deng, Jiankang and Guo, Jia and Yang, Jing and Xue, Niannan and Kotsia, Irene and Zafeiriou, Stefanos},
	journal={IEEE Transactions on Pattern Analysis and Machine Intelligence}, 
	title={ArcFace: Additive Angular Margin Loss for Deep Face Recognition}, 
	year={2022},
	volume={44},
	number={10},
	pages={5962-5979}}

@INPROCEEDINGS{huang2020curricularface,
	author={Huang, Yuge and Wang, Yuhan and Tai, Ying and Liu, Xiaoming and Shen, Pengcheng and Li, Shaoxin and Li, Jilin and Huang, Feiyue},
	booktitle={2020 IEEE/CVF Conference on Computer Vision and Pattern Recognition (CVPR)}, 
	title={CurricularFace: Adaptive Curriculum Learning Loss for Deep Face Recognition}, 
	year={2020},
	volume={},
	number={},
	pages={5900-5909}}

@inproceedings{wang2020mis,
	title={Mis-classified vector guided softmax loss for face recognition},
	author={Wang, Xiaobo and Zhang, Shifeng and Wang, Shuo and Fu, Tianyu and Shi, Hailin and Mei, Tao},
	booktitle={Proceedings of the AAAI Conference on Artificial Intelligence},
	volume={34},
	number={07},
	pages={12241--12248},
	year={2020}
}

@INPROCEEDINGS{shi2019probabilistic,
	author={Shi, Yichun and Jain, Anil},
	booktitle={2019 IEEE/CVF International Conference on Computer Vision (ICCV)}, 
	title={Probabilistic Face Embeddings}, 
	year={2019},
	volume={},
	number={},
	pages={6901-6910}}

@INPROCEEDINGS{ijbb,
	author={Whitelam, Cameron and Taborsky, Emma and Blanton, Austin and Maze, Brianna and Adams, Jocelyn and Miller, Tim and Kalka, Nathan and Jain, Anil K. and Duncan, James A. and Allen, Kristen and Cheney, Jordan and Grother, Patrick},
	booktitle={2017 IEEE Conference on Computer Vision and Pattern Recognition Workshops (CVPRW)}, 
	title={IARPA Janus Benchmark-B Face Dataset}, 
	year={2017},
	volume={},
	number={},
	pages={592-600}}

@INPROCEEDINGS{ijbc,
	author={Maze, Brianna and Adams, Jocelyn and Duncan, James A. and Kalka, Nathan and Miller, Tim and Otto, Charles and Jain, Anil K. and Niggel, W. Tyler and Anderson, Janet and Cheney, Jordan and Grother, Patrick},
	booktitle={2018 International Conference on Biometrics (ICB)}, 
	title={IARPA Janus Benchmark - C: Face Dataset and Protocol}, 
	year={2018},
	volume={},
	number={},
	pages={158-165}}

@inproceedings{lfw,
	title={Labeled faces in the wild: A database for studying face recognition in unconstrained environments},
	author={Huang, Gary B and Mattar, Marwan and Berg, Tamara and Learned-Miller, Eric},
	booktitle={Workshop on faces in'Real-Life'Images: detection, alignment, and recognition},
	year={2008}
}

@INPROCEEDINGS{cfpfp,
	author={Sengupta, Soumyadip and Chen, Jun-Cheng and Castillo, Carlos and Patel, Vishal M. and Chellappa, Rama and Jacobs, David W.},
	booktitle={2016 IEEE Winter Conference on Applications of Computer Vision (WACV)}, 
	title={Frontal to profile face verification in the wild}, 
	year={2016},
	volume={},
	number={},
	pages={1-9}}

@article{cplfw,
	title={Cross-pose lfw: A database for studying cross-pose face recognition in unconstrained environments},
	author={Zheng, Tianyue and Deng, Weihong},
	journal={Beijing University of Posts and Telecommunications, Tech. Rep},
	volume={5},
	number={7},
	year={2018}
}

@INPROCEEDINGS{agedb,
	author={Moschoglou, Stylianos and Papaioannou, Athanasios and Sagonas, Christos and Deng, Jiankang and Kotsia, Irene and Zafeiriou, Stefanos},
	booktitle={2017 IEEE Conference on Computer Vision and Pattern Recognition Workshops (CVPRW)}, 
	title={AgeDB: The First Manually Collected, In-the-Wild Age Database}, 
	year={2017},
	volume={},
	number={},
	pages={1997-2005}}

@article{calfw,
	title={Cross-age lfw: A database for studying cross-age face recognition in unconstrained environments},
	author={Zheng, Tianyue and Deng, Weihong and Hu, Jiani},
	journal={arXiv preprint arXiv:1708.08197},
	year={2017}
}

@INPROCEEDINGS{he2016deep,
	author={He, Kaiming and Zhang, Xiangyu and Ren, Shaoqing and Sun, Jian},
	booktitle={2016 IEEE Conference on Computer Vision and Pattern Recognition (CVPR)}, 
	title={Deep Residual Learning for Image Recognition}, 
	year={2016},
	volume={},
	number={},
	pages={770-778}}

@INPROCEEDINGS{li2021spherical,
	author={Li, Shen and Xu, Jianqing and Xu, Xiaqing and Shen, Pengcheng and Li, Shaoxin and Hooi, Bryan},
	booktitle={2021 IEEE/CVF Conference on Computer Vision and Pattern Recognition (CVPR)}, 
	title={Spherical Confidence Learning for Face Recognition}, 
	year={2021},
	volume={},
	number={},
	pages={15624-15632}}

@inproceedings{tinyface,
	title={Low-resolution face recognition},
	author={Cheng, Zhiyi and Zhu, Xiatian and Gong, Shaogang},
	booktitle={Computer Vision--ACCV 2018: 14th Asian Conference on Computer Vision, Perth, Australia, December 2--6, 2018, Revised Selected Papers, Part III 14},
	pages={605--621},
	year={2019},
}

@INPROCEEDINGS{meng2021magface,
	author={Meng, Qiang and Zhao, Shichao and Huang, Zhida and Zhou, Feng},
	booktitle={2021 IEEE/CVF Conference on Computer Vision and Pattern Recognition (CVPR)}, 
	title={MagFace: A Universal Representation for Face Recognition and Quality Assessment}, 
	year={2021},
	volume={},
	number={},
	pages={14220-14229}}

@INPROCEEDINGS{kim2022adaface,
	author={Kim, Minchul and Jain, Anil K. and Liu, Xiaoming},
	booktitle={2022 IEEE/CVF Conference on Computer Vision and Pattern Recognition (CVPR)}, 
	title={AdaFace: Quality Adaptive Margin for Face Recognition}, 
	year={2022},
	volume={},
	number={},
	pages={18729-18738}}

@INPROCEEDINGS{vgg2fp,
	author={Cao, Qiong and Shen, Li and Xie, Weidi and Parkhi, Omkar M. and Zisserman, Andrew},
	booktitle={2018 13th IEEE International Conference on Automatic Face \& Gesture Recognition (FG 2018)}, 
	title={VGGFace2: A Dataset for Recognising Faces across Pose and Age}, 
	year={2018},
	volume={},
	number={},
	pages={67-74}}

@INPROCEEDINGS{vlrfr2,
	author={Long Chai, Jacky Chen and Ng, Tiong-Sik and Low, Cheng-Yaw and Park, Jaewoo and Jin Teoh, Andrew Beng},
	booktitle={2023 IEEE/CVF Conference on Computer Vision and Pattern Recognition (CVPR)}, 
	title={Recognizability Embedding Enhancement for Very Low-Resolution Face Recognition and Quality Estimation}, 
	year={2023},
	volume={},
	number={},
	pages={9957-9967}}

@INPROCEEDINGS{3d-berl,
	author={He, Mingjie and Zhang, Jie and Shan, Shiguang and Chen, Xilin},
	booktitle={2022 IEEE/CVF Conference on Computer Vision and Pattern Recognition (CVPR)}, 
	title={Enhancing Face Recognition with Self-Supervised 3D Reconstruction}, 
	year={2022},
	volume={},
	number={},
	pages={4052-4061}}

@INPROCEEDINGS{liu2022diverse_data,
	author={Liu, Chang and Yu, Xiang and Tsai, Yi-Hsuan and Faraki, Masoud and Moslemi, Ramin and Chandraker, Manmohan and Fu, Yun},
	booktitle={2022 IEEE/CVF Conference on Computer Vision and Pattern Recognition (CVPR)}, 
	title={Learning to Learn across Diverse Data Biases in Deep Face Recognition}, 
	year={2022},
	volume={},
	number={},
	pages={4062-4072}}

@ARTICLE{spherefaceRV,
	author={Liu, Weiyang and Wen, Yandong and Raj, Bhiksha and Singh, Rita and Weller, Adrian},
	journal={IEEE Transactions on Pattern Analysis and Machine Intelligence}, 
	title={SphereFace Revived: Unifying Hyperspherical Face Recognition}, 
	year={2023},
	volume={45},
	number={2},
	pages={2458-2474}}

@INPROCEEDINGS{contrastiveloss,
	author={Hadsell, R. and Chopra, S. and LeCun, Y.},
	booktitle={IEEE Computer Society Conference on Computer Vision and Pattern Recognition (CVPR)}, 
	title={Dimensionality Reduction by Learning an Invariant Mapping}, 
	year={2006},
	number={},
	pages={1735-1742}}

@INPROCEEDINGS{facenet,
	author={Schroff, Florian and Kalenichenko, Dmitry and Philbin, James},
	booktitle={IEEE Conference on Computer Vision and Pattern Recognition (CVPR)}, 
	title={FaceNet: A unified embedding for face recognition and clustering}, 
	year={2015},
	volume={},
	number={},
	pages={815-823}}

@INPROCEEDINGS{Lifted_Structure_Loss,
	author={Song, Hyun Oh and Xiang, Yu and Jegelka, Stefanie and Savarese, Silvio},
	booktitle={IEEE Conference on Computer Vision and Pattern Recognition (CVPR)}, 
	title={Deep Metric Learning via Lifted Structured Feature Embedding}, 
	year={2016},
	volume={},
	number={},
	pages={4004-4012}}

@INPROCEEDINGS{Binomial_Deviance_Loss,
	author={Opitz, Michael and Waltner, Georg and Possegger, Horst and Bischof, Horst},
	booktitle={IEEE International Conference on Computer Vision (ICCV)}, 
	title={BIER — Boosting Independent Embeddings Robustly}, 
	year={2017},
	volume={},
	number={},
	pages={5199-5208}}

@ARTICLE{ucface,
	author={Ahn, Kyeongjin and Lee, Seungeon and Han, Sungwon and Low, Cheng Yaw and Cha, Meeyoung},
	journal={IEEE Transactions on Information Forensics and Security}, 
	title={Uncertainty-Aware Face Embedding With Contrastive Learning for Open-Set Evaluation}, 
	year={2024},
	volume={19},
	number={},
	pages={7176-7186}}

@ARTICLE{unsup3d,
	author={Wu, Shangzhe and Rupprecht, Christian and Vedaldi, Andrea},
	journal={IEEE Transactions on Pattern Analysis and Machine Intelligence}, 
	title={Unsupervised Learning of Probably Symmetric Deformable 3D Objects From Images in the Wild (Invited Paper)}, 
	year={2023},
	volume={45},
	number={4},
	pages={5268-5281}}

@article{FRABSM,
	title = {Robust face recognition model based sample mining and loss functions},
	journal = {Knowledge-Based Systems},
	volume = {302},
	pages = {112330},
	year = {2024},
	issn = {0950-7051},
	author = {Yang Wang and Fan Xie and Chuanxin Zhao and Ao Wang and Chang Ma and Shijia Song and Zhenyu Yuan and Lijun Zhao},
}

@INPROCEEDINGS{dcface,
	author={Kim, Minchul and Liu, Feng and Jain, Anil and Liu, Xiaoming},
	booktitle={2023 IEEE/CVF Conference on Computer Vision and Pattern Recognition (CVPR)}, 
	title={DCFace: Synthetic Face Generation with Dual Condition Diffusion Model}, 
	year={2023},
	volume={},
	number={},
	pages={12715-12725}}

@INPROCEEDINGS{refo,
	author={Li, Jingzhi and Guo, Zidong and Li, Hui and Han, Seungju and Baek, Ji-Won and Yang, Min and Yang, Ran and Suh, Sungjoo},
	booktitle={2023 IEEE/CVF Conference on Computer Vision and Pattern Recognition (CVPR)}, 
	title={Rethinking Feature-based Knowledge Distillation for Face Recognition}, 
	year={2023},
	volume={},
	number={},
	pages={20156-20165}}

@inproceedings{iic,
	title={Enhanced Face Recognition using Intra-class Incoherence Constraint},
	author={Yuanqing Huang and Yinggui Wang and Le Yang and Lei Wang},
	booktitle={The Twelfth International Conference on Learning Representations},
	year={2024}
}

@INPROCEEDINGS{kprpe,
	author={Kim, Minchul and Su, Yiyang and Liu, Feng and Jain, Anil and Liu, Xiaoming},
	booktitle={2024 IEEE/CVF Conference on Computer Vision and Pattern Recognition (CVPR)}, 
	title={KeyPoint Relative Position Encoding for Face Recognition}, 
	year={2024},
	volume={},
	number={},
	pages={244-255}}

@INPROCEEDINGS{lafs,
	author={Sun, Zhonglin and Feng, Chen and Patras, Ioannis and Tzimiropoulos, Georgios},
	booktitle={2024 IEEE/CVF Conference on Computer Vision and Pattern Recognition (CVPR)}, 
	title={LAFS: Landmark-Based Facial Self-Supervised Learning for Face Recognition}, 
	year={2024},
	volume={},
	number={},
	pages={1639-1649}}

@ARTICLE{qsd,
	author={Ou, Fu-Zhao and Chen, Xingyu and Zhao, Kai and Wang, Shiqi and Wang, Yuan-Gen and Kwong, Sam},
	journal={IEEE Transactions on Multimedia}, 
	title={Refining Uncertain Features With Self-Distillation for Face Recognition and Person Re-Identification}, 
	year={2024},
	volume={26},
	number={},
	pages={6981-6995}}

@article{coreface,
	title = {CoReFace: Sample-guided Contrastive Regularization for Deep Face Recognition},
	journal = {Pattern Recognition},
	volume = {152},
	pages = {110483},
	year = {2024},
	issn = {0031-3203},
	author = {Youzhe Song and Feng Wang},
}

@ARTICLE{joint-berl,
	author={He, Mingjie and Zhang, Jie and Shan, Shiguang and Chen, Xilin},
	journal={IEEE Transactions on Image Processing}, 
	title={Enhancing Face Recognition With Detachable Self-Supervised Bypass Networks}, 
	year={2024},
	volume={33},
	number={},
	pages={1588-1599}}

@ARTICLE{id-gan,
	author={Ge, Shiming and Li, Chenyu and Zhao, Shengwei and Zeng, Dan},
	journal={IEEE Transactions on Circuits and Systems for Video Technology}, 
	title={Occluded Face Recognition in the Wild by Identity-Diversity Inpainting}, 
	year={2020},
	volume={30},
	number={10},
	pages={3387-3397}}

@INPROCEEDINGS{face-cycle,
	author={Chang, Jia-Ren and Chen, Yong-Sheng and Chiu, Wei-Chen},
	booktitle={2021 IEEE/CVF International Conference on Computer Vision (ICCV)}, 
	title={Learning Facial Representations from the Cycle-consistency of Face}, 
	year={2021},
	volume={},
	number={},
	pages={9660-9669}}

@INPROCEEDINGS{lap,
	author={Zhang, Zhenyu and Ge, Yanhao and Chen, Renwang and Tai, Ying and Yan, Yan and Yang, Jian and Wang, Chengjie and Li, Jilin and Huang, Feiyue},
	booktitle={2021 IEEE/CVF Conference on Computer Vision and Pattern Recognition (CVPR)}, 
	title={Learning to Aggregate and Personalize 3D Face from In-the-Wild Photo Collection}, 
	year={2021},
	volume={},
	number={},
	pages={14209-14219}}

@ARTICLE{mdfr,
	author={Tu, Xiaoguang and Zhao, Jian and Liu, Qiankun and Ai, Wenjie and Guo, Guodong and Li, Zhifeng and Liu, Wei and Feng, Jiashi},
	journal={IEEE Transactions on Circuits and Systems for Video Technology}, 
	title={Joint Face Image Restoration and Frontalization for Recognition}, 
	year={2022},
	volume={32},
	number={3},
	pages={1285-1298}}

@INPROCEEDINGS{l2r,
	author={Zhang, Zhenyu and Ge, Yanhao and Tai, Ying and Huang, Xiaoming and Wang, Chengjie and Tang, Hao and Huang, Dongjin and Xie, Zhifeng},
	booktitle={2022 IEEE/CVF Conference on Computer Vision and Pattern Recognition (CVPR)}, 
	title={Learning to Restore 3D Face from In-the-Wild Degraded Images}, 
	year={2022},
	volume={},
	number={},
	pages={4227-4237}}

@INPROCEEDINGS{phydir,
	author={Zhang, Zhenyu and Ge, Yanhao and Tai, Ying and Cao, Weijian and Chen, Renwang and Liu, Kunlin and Tang, Hao and Huang, Xiaoming and Wang, Chengjie and Xie, Zhifeng and Huang, Dongjin},
	booktitle={2022 IEEE/CVF Conference on Computer Vision and Pattern Recognition (CVPR)}, 
	title={Physically-guided Disentangled Implicit Rendering for 3D Face Modeling}, 
	year={2022},
	volume={},
	number={},
	pages={20321-20331}}

@INPROCEEDINGS{igc-net,
	author={Yu, Chang and Zhu, Xiangyu and Zhang, Xiaomei and Zhang, Zhaoxiang and Lei, Zhen},
	booktitle={2023 IEEE/CVF Conference on Computer Vision and Pattern Recognition (CVPR)}, 
	title={Graphics Capsule: Learning Hierarchical 3D Face Representations from 2D Images}, 
	year={2023},
	volume={},
	number={},
	pages={20981-20990}}

@INPROCEEDINGS{celeba,
	author={Liu, Ziwei and Luo, Ping and Wang, Xiaogang and Tang, Xiaoou},
	booktitle={2015 IEEE International Conference on Computer Vision (ICCV)}, 
	title={Deep Learning Face Attributes in the Wild}, 
	year={2015},
	volume={},
	number={},
	pages={3730-3738}}

@ARTICLE{frl,
	author={Xin, Jingwei and Wei, Zikai and Wang, Nannan and Li, Jie and Gao, Xinbo},
	journal={IEEE Transactions on Information Forensics and Security}, 
	title={Large Pose Face Recognition via Facial Representation Learning}, 
	year={2024},
	volume={19},
	number={},
	pages={934-946}}

@ARTICLE{virface,
	author={Li, Wenyu and Li, Pengyu and Guo, Tianchu and Chen, Binghui and Wang, Biao and Zuo, Wangmeng and Zhang, Lei},
	journal={IEEE Transactions on Information Forensics and Security}, 
	title={VirFace$^\infty$: A Semi-Supervised Method for Enhancing Face Recognition via Unlabeled Shallow Data}, 
	year={2023},
	volume={18},
	number={},
	pages={5148-5159}}

@ARTICLE{ddc,
	author={Uzun, Bedirhan and Cevikalp, Hakan and Saribas, Hasan},
	journal={IEEE Transactions on Pattern Analysis and Machine Intelligence}, 
	title={Deep Discriminative Feature Models (DDFMs) for Set Based Face Recognition and Distance Metric Learning}, 
	year={2023},
	volume={45},
	number={5},
	pages={5594-5608}}

@ARTICLE{facet-b,
	author={Zhu, Yuhao and Ren, Min and Jing, Hui and Dai, Linlin and Sun, Zhenan and Li, Ping},
	journal={IEEE Transactions on Information Forensics and Security}, 
	title={Joint Holistic and Masked Face Recognition}, 
	year={2023},
	volume={18},
	number={},
	pages={3388-3400}}
\begin{IEEEbiography}[{\includegraphics[width=1in,height=1.25in,clip,keepaspectratio]{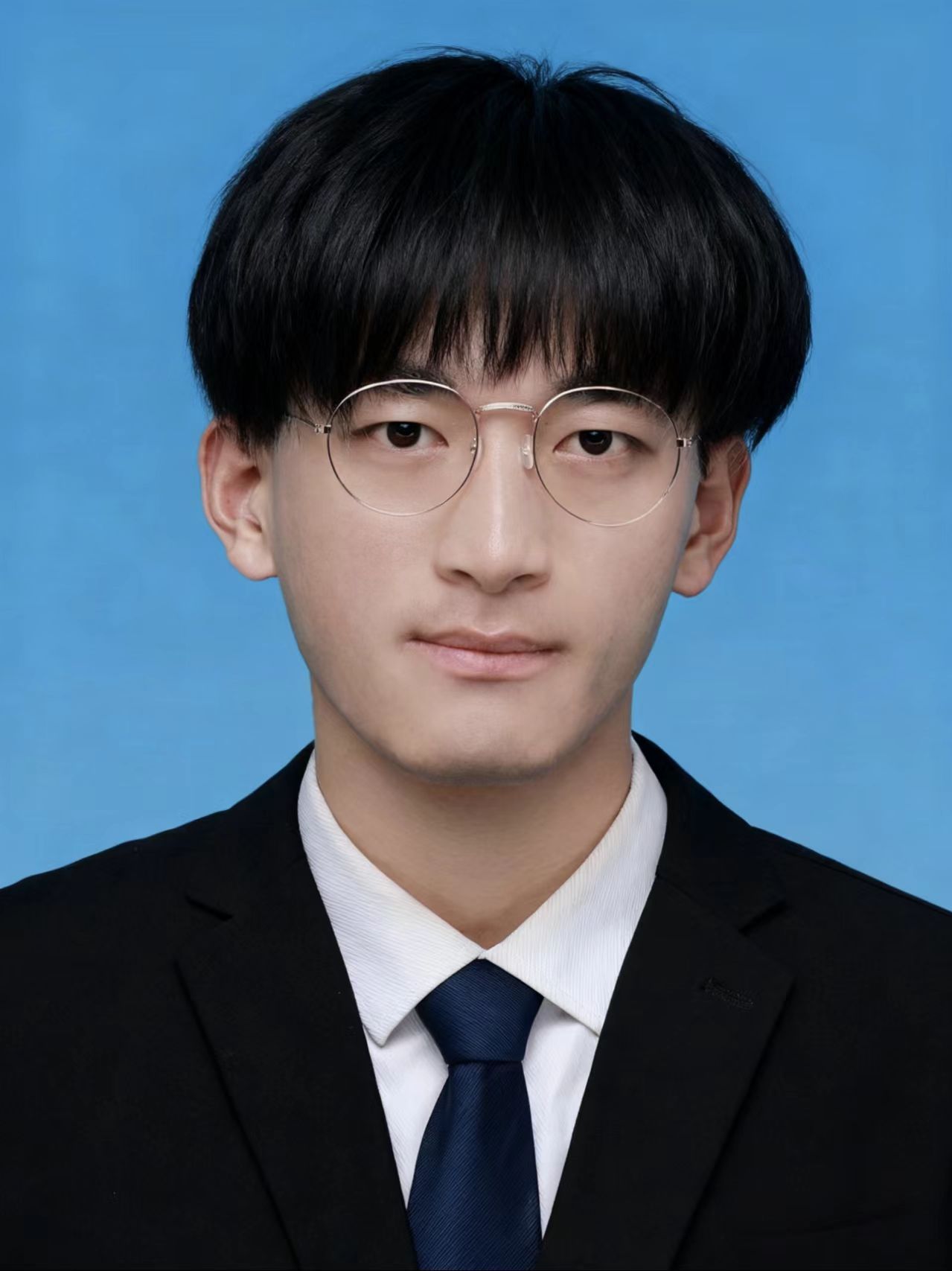}}]{Fan Xie}
	received the M.Sc. degree with the School of Computer and
	Information, Anhui Normal University, Wuhu, China, in 2025. He is currently an algorithm engineer. His research interests include face recognition, face reconstruction, robot control system, and convex optimization.
\end{IEEEbiography}
\begin{IEEEbiography}[{\includegraphics[width=1in,height=1.25in,clip,keepaspectratio]{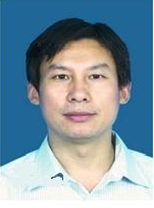}}]{Yang Wang}
	(Member, IEEE) is a Professor at Anhui Normal University now. He received his Bachelor’s degree from Anhui Normal University in 1994, MS degree from China West Normal University in 2005, and PhD degree from Soochow University in 2009. He is a Postdoctoral Fellow between Dec. 2009 and Jul. 2011 at the University of Science and Technology of China. He has published more than 80 papers. His current research interests include computer vision, augmented reality, Internet of Vehicles and information system optimization.
\end{IEEEbiography}
\begin{IEEEbiography}[{\includegraphics[width=1in,height=1.25in,clip,keepaspectratio]{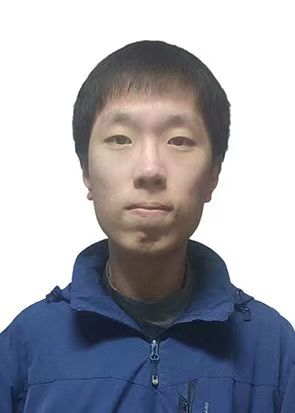}}]{Yikang Jiao}
	is currently pursuing the M.S. degree with Anhui Normal University, Wuhu, China. His research interests include image processing, computer vision and machine learning.
\end{IEEEbiography}
\begin{IEEEbiography}[{\includegraphics[width=1in,height=1.25in,clip,keepaspectratio]{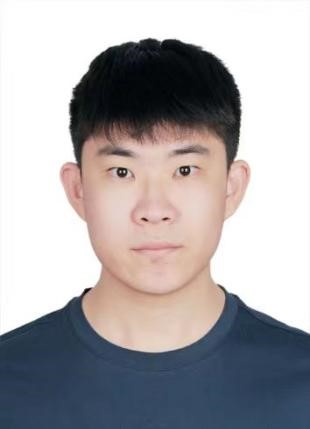}}]{Zhenyu Yuan}
	is currently pursuing the M.S. degree with Anhui Normal University, Wuhu, China. His research interests include image processing, computer vision, ecological informatics, etc.
\end{IEEEbiography}
\begin{IEEEbiography}[{\includegraphics[width=1in,height=1.25in,clip,keepaspectratio]{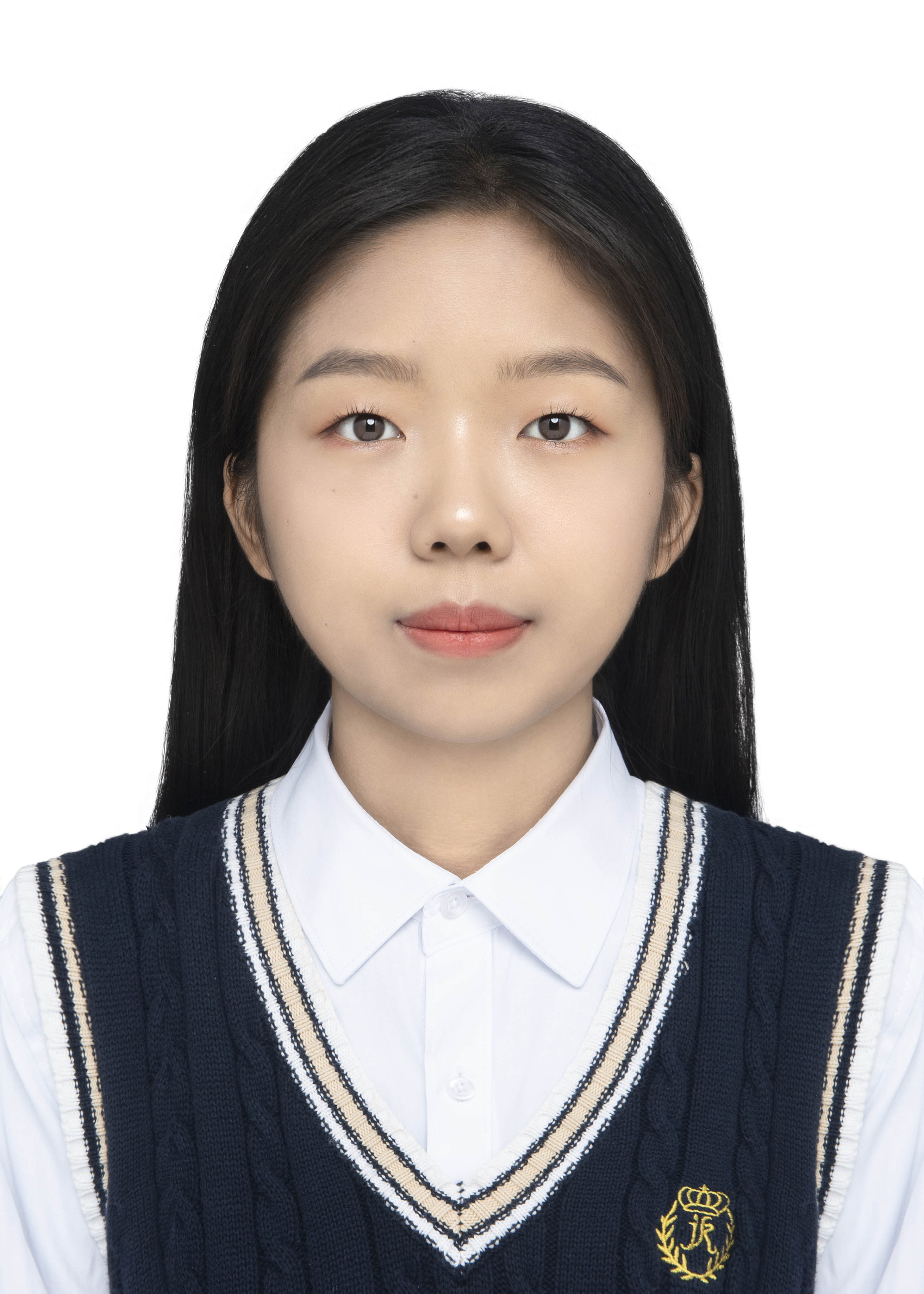}}]{Congxi Chen}
	is currently pursuing the M.S. degree with Anhui Normal University, Wuhu, China. Her research interests include artificial intelligence, computer vision, etc.
\end{IEEEbiography}
\begin{IEEEbiography}[{\includegraphics[width=1in,height=1.25in,clip,keepaspectratio]{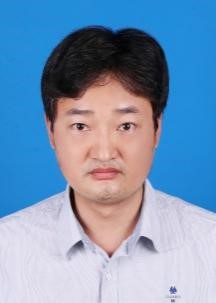}}]{Chuanxin Zhao}
	(Member, IEEE) is a Professor at Anhui Normal University. He received his Ph.D. in computer science from Soochow University, Suzhou, China, in 2013. He was a visitor Research Fellow in Curtin University between Aug. 2014 and Jul. 2015. He has published more than 50 papers and applied 10 patents. His current research interests include IoT, privacy protection, information security.
\end{IEEEbiography}

\vfill

\clearpage
\appendices

\section{Speed Improvement and Quality Optimization of Depth Map Rotation}
In this section, we propose an efficient depth map rotation and mapping method, aiming to address the issues of low computational efficiency and poor quality in traditional methods when handling depth maps. By introducing vectorized operations and advanced indexing techniques, improvements are successfully achieved in both the speed and quality of depth map rotation.

\subsection{Method Overview}
A depth map rotation method is introduced based on vectorized operations. The method mainly includes the following key steps:

\subsubsection{Conversion from Depth Map to 3D Point Cloud}
First, the original depth map $\mathbf{d}$ is converted into a 3D point cloud $\bm{P}$ in the camera coordinate system. For each pixel coordinate $(u, v)$ in the depth map and its corresponding depth value $\mathbf{d}(u, v)$, the corresponding 3D coordinates are computed using the camera intrinsic matrix $\mathbf{K}$ and its inverse matrix $\mathbf{K}^{-1}$ as follows:
\begin{equation}
	\begin{bmatrix}
		x \\
		y \\
		z \\
	\end{bmatrix}
	= \mathbf{d}(u, v) \cdot \mathbf{K}^{-1} \begin{bmatrix}
		u \\
		v \\
		1 \\
	\end{bmatrix}
\end{equation}

\subsubsection{Rotation and Translation of 3D Point Cloud}
The 3D point cloud $\bm{P}$ is transformed using a rotation matrix $\mathbf{R}$ and a translation vector $\bm{t}$ as follows:
\begin{equation}
	\bm{P}' = \mathbf{R} \cdot \bm{P} + \bm{t}
\end{equation}
where the rotation center is chosen to be the center of the scene's depth in order to maintain smoothness and consistency in the transformation.

\subsubsection{Projection of the 3D Point Cloud Back to the New Depth Map}
The transformed 3D point cloud $P'$ is projected back onto the new image plane to obtain the new pixel coordinates $(u', v')$ and depth values $d'$:
\begin{equation}
	\begin{bmatrix}
		u' \\
		v' \\
		w \\
	\end{bmatrix}
	= \mathbf{K} \cdot \bm{P}'
	\quad \Rightarrow \quad
	u' = \frac{u'}{w}, \quad v' = \frac{v'}{w}, \quad d' = w
\end{equation}

\subsubsection{Vectorized Depth Map Update}
When mapping the transformed depth values back to the new depth map, methods that rely on nested loops to update each pixel and its neighbors one by one result in a large computational cost and fail to fully leverage the parallel computing capabilities of modern hardware. To address it, we introduce a vectorized depth map update method, utilizing advanced indexing and scatter-reduction operations to efficiently map the new depth values back into the new depth map, properly handle occlusion relationships, and avoid the computational overhead of nested loops.

\subsection{Technical Details}

\subsubsection{Normalization and Discretization of Coordinate Indices}
To map continuous pixel coordinates to discrete pixel indices, normalization and scaling of the coordinates are performed:
\begin{equation}
	\begin{split}
		&i = \left\lfloor \frac{(u' - u_{\min}) \cdot (W_{\text{new}} - 1)}{u_{\max} - u_{\min}} + 0.5 \right\rfloor \\ 
		&j = \left\lfloor \frac{(v' - v_{\min}) \cdot (H_{\text{new}} - 1)}{v_{\max} - v_{\min}} + 0.5 \right\rfloor 
	\end{split}
\end{equation}
where $(u', v')$ are the projected continuous pixel coordinates, $(i, j)$ are the corresponding integer pixel indices, and $W_{\text{new}}$ and $H_{\text{new}}$ are the new depth map width and height. The addition of $0.5$ ensures rounding to the nearest integer.

\subsubsection{Generation of Neighborhood Offsets}
To perform neighborhood depth updates within a specified radius, we pre-generate a list of neighborhood offsets $(\Delta i, \Delta j)$ that satisfy the following condition:
\begin{equation}
	\Delta i^2 + \Delta j^2 \leq r^2
\end{equation}
where $r$ is the neighborhood radius.

\subsubsection{Advanced Indexing and Scatter Reduction Operation}
Using PyTorch's `scatter\_reduce` function, we efficiently updated the depth map. This function allows for reduction operations at specified locations even in the presence of duplicate indices. The 'amin' reduction was chosen to minimize the values. This ensures that in overlapping regions, the smallest depth value (i.e., the nearest object) is retained, thus correctly handling occlusion relationships.

Specifically, we first flatten the 3D batch, row, and column indices $(n, i, j)$ into a 1D linear index:
\begin{equation}
	l = n \cdot (H_{\text{new}} \cdot W_{\text{new}}) + i \cdot W_{\text{new}} + j
\end{equation}
Then, we perform the scatter reduction operation on the flattened depth map vector $\bm{d}_{\text{flat}}$:
\begin{equation}
	\bm{d}_{\text{flat}}[l] = \min\left( \bm{d}_{\text{flat}}[l], d_{\text{new}} \right)
\end{equation}
where $d_{\text{new}}$ is the depth value to be updated.

\subsubsection{Advantages of Vectorization}
Through the methods described above, we transformed the depth map update operation, which originally requires nested loops, into tensor-based vectorized computation. This not only significantly improved computational efficiency but also fully leveraged the parallel computing power of GPUs. At the same time, using scatter reduction operations to handle duplicate indices ensures the correctness of depth value updates.

\begin{figure}[h]
	\begin{center}
		\includegraphics[width=1.0\linewidth]{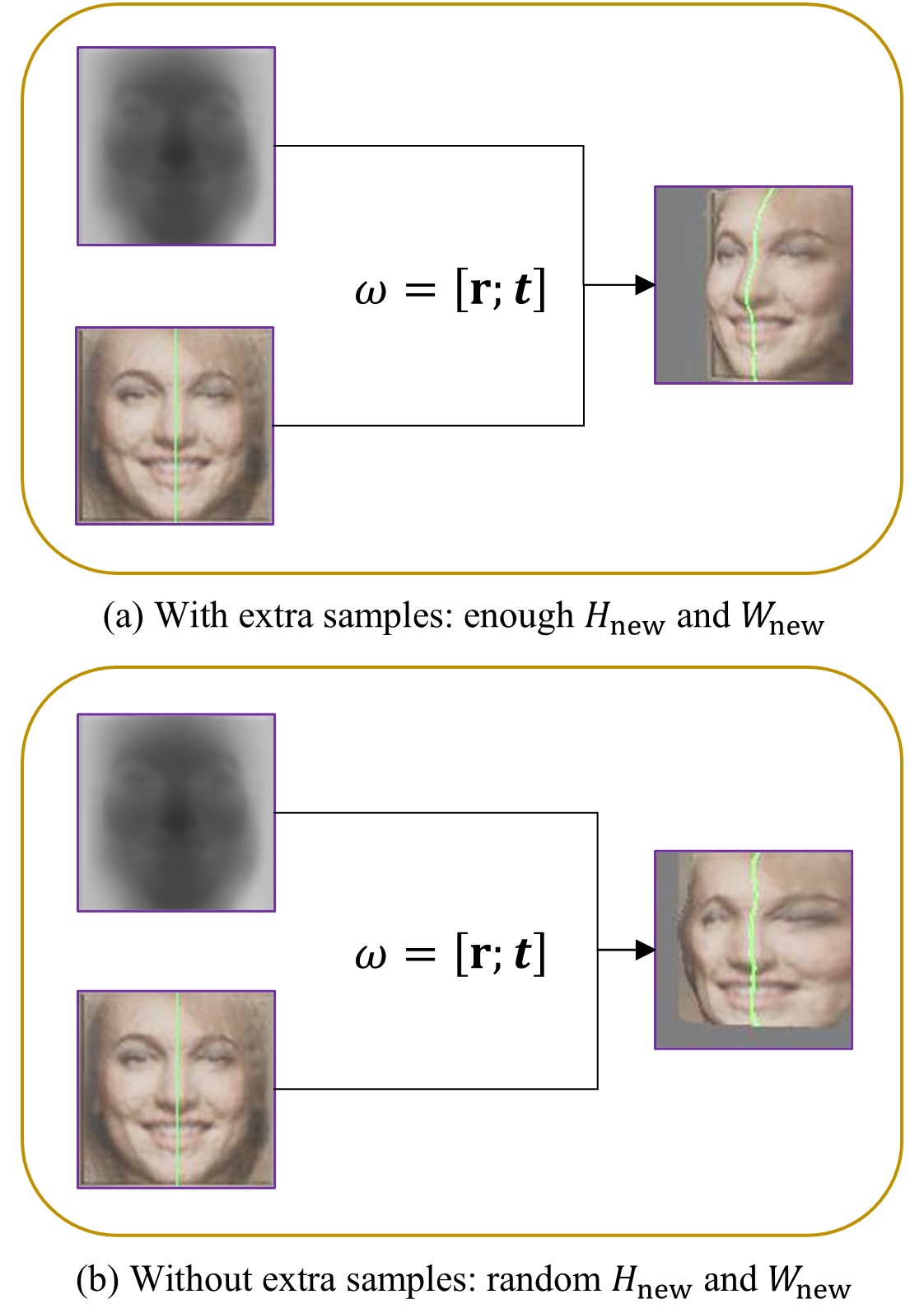}
	\end{center}
	\caption{The image shows the changes in the distorted image before and after using enough $H_\text{new}$ and $W_\text{new}$. It can be seen that our method produces the correct distortion results.}
	\label{figure_renderer}
\end{figure}

\begin{figure*}[h]
	\begin{center}
		\includegraphics[width=1.0\linewidth]{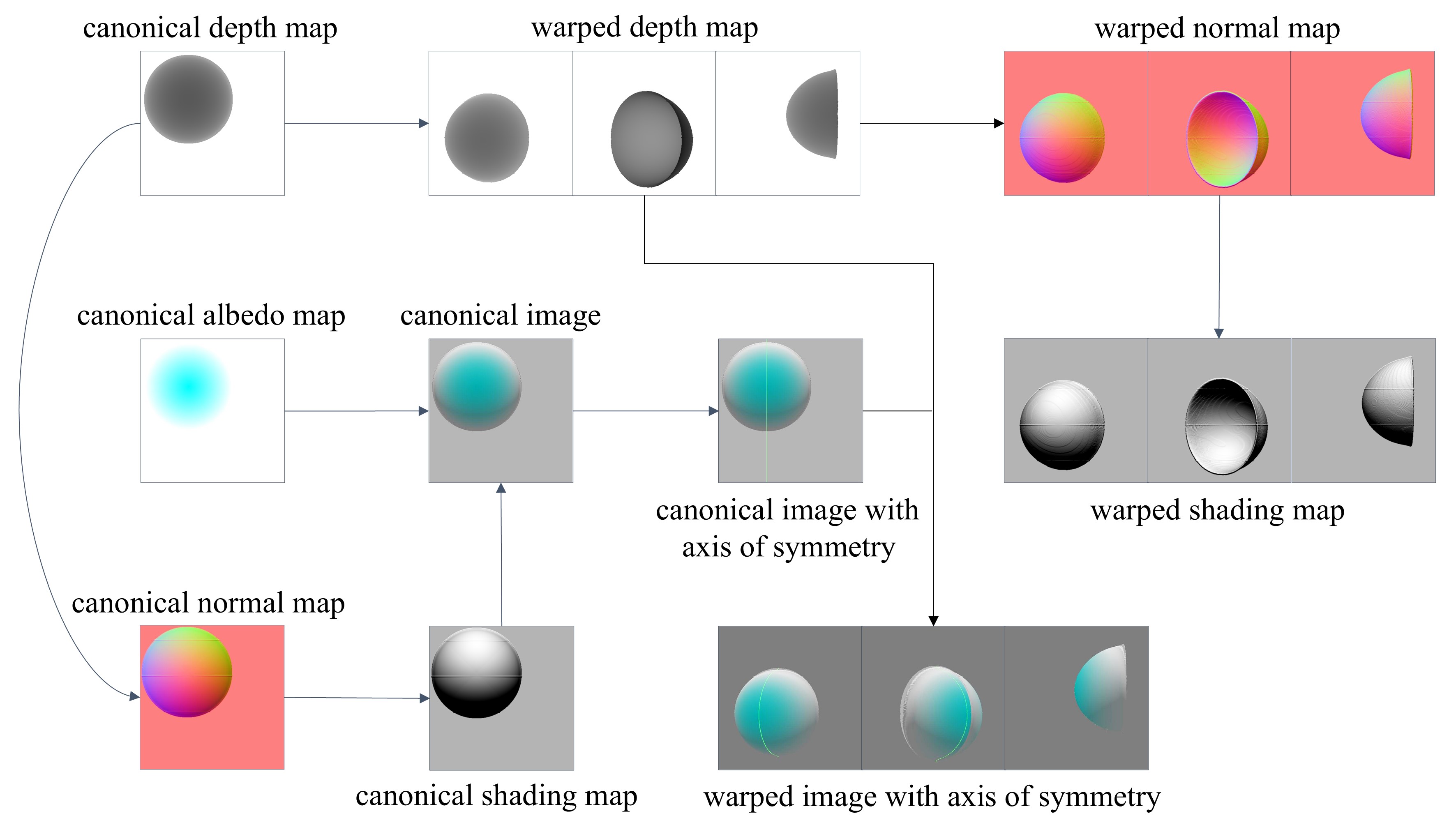}
	\end{center}
	\caption{The image shows a toy example of our renderer. It is the result of a hemisphere rotating in three different directions. A symmetry axis was manually added to the hemisphere. The lighting parameters of the hemisphere remain unchanged during the rotation.}
	\label{renderer_ex}
\end{figure*}

\subsection{Uniform Depth Map Size to Eliminate Alignment Issues}
When performing depth map rotation and remapping, the rotation angles and displacements of different images vary during the transformation process, resulting in inconsistent sizes for the transformed depth maps (i.e., $H_\text{new}$ and $W_\text{new}$). This size inconsistency can lead to severe alignment issues when generating side-view RGB images from the depth maps, causing distorted RGB image outputs. To address this problem, a uniform size strategy is proposed. The core idea is to introduce two additional auxiliary images as extra boundaries, expanding the $H_\text{new}$ and $W_\text{new}$ of all images and maintaining symmetry in the expansion.

\subsubsection{Expansion Strategy}

\paragraph{Necessity of Uniform Size}
In the rotated depth maps, the values of $x_{\min}$, $x_{\max}$, $y_{\min}$, and $y_{\max}$ differ due to varying rotation angles. For example, suppose the original depth map has a size of $H \times W$. After rotation, for the $i$-th depth map, it may happen that: $x_{\min}^{(i)} = 12.7, \, x_{\max}^{(i)} = 145.15, \, y_{\min}^{(i)} = -84.29, \, y_{\max}^{(i)} = 96.2$. The resulting $H_\text{new}^{(i)} = 180.49$ and $W_\text{new}^{(i)} = 157.85$ are different from those of other images. This discrepancy will lead to alignment issues in subsequent processing.

\paragraph{Introduction of Auxiliary Images}
To address this issue, two auxiliary images are introduced specifically to expand the dimensions of $H_\text{new}$ and $W_\text{new}$. These two auxiliary images do not participate in subsequent loss calculations or image generation; they are solely used to determine a unified depth map size.

\paragraph{Computation of the Expansion}
When determining the unified $H_\text{new}$ and $W_\text{new}$, the first step is to compute the global range of $x_{\min}$, $x_{\max}$, $y_{\min}$, and $y_{\max}$ across all images:
\begin{equation}
	\begin{split}
		&x_{\min}^{\text{global}} = \min(x_{\min}^{(1)}, x_{\min}^{(2)}, \dots, x_{\min}^{(N)}) \\ 
		&x_{\max}^{\text{global}} = \max(x_{\max}^{(1)}, x_{\max}^{(2)}, \dots, x_{\max}^{(N)}) \\
		&y_{\min}^{\text{global}} = \min(y_{\min}^{(1)}, y_{\min}^{(2)}, \dots, y_{\min}^{(N)}) \\
		&y_{\max}^{\text{global}} = \max(y_{\max}^{(1)}, y_{\max}^{(2)}, \dots, y_{\max}^{(N)})
	\end{split}
\end{equation}
where $(N + 2)$ represents the total number of images, including the two auxiliary images.

By calculating these global extrema, the unified dimensions can be determined as:
\begin{equation}
	\begin{split}
		&H_\text{new}^{\text{global}} = \lceil y_{\max}^{\text{global}} - y_{\min}^{\text{global}} \rceil \\
		&W_\text{new}^{\text{global}} = \lceil x_{\max}^{\text{global}} - x_{\min}^{\text{global}} \rceil
	\end{split}
\end{equation}

\paragraph{Implementation of Symmetric Expansion}
To maintain the center alignment of the transformed images, we always use the center of the original image as a reference during the expansion process, ensuring that the expansion ranges on both sides are symmetric and preventing the introduction of additional offsets. Therefore, the following conditions must be ensured:
\begin{equation}
	\begin{split}
		&x_{\min}^{\text{global}} + x_{\max}^{\text{global}} = W \\
		&y_{\min}^{\text{global}} + y_{\max}^{\text{global}} = H \\
		&W_\text{new}^{\text{global}} / H_\text{new}^{\text{global}} = W / H
	\end{split}
\end{equation}
where $W_\text{new}^{\text{global}} / W$ needs to be sufficiently large, approximately above $2.5$.

\subsubsection{Cropping Operation and Final Alignment}

\paragraph{Cropping the Depth Map}  
After generating the new depth map, it is immediately cropped to the center region of the original size $H \times W$:
\begin{equation}
	\text{Crop}_{\text{center}}(\mathbf{d}, H, W)
\end{equation}
This cropping operation helps reduce memory and computational overhead while ensuring that the center region of the depth map is aligned with the original image.

\paragraph{Final Alignment with RGB Images}  
When mapping the side depth map to an RGB image using the `nn.functional.grid\_sample` function, the cropped depth map is fully aligned, preventing distortion issues in the RGB image caused by size inconsistencies.

\subsubsection{Comparison and Advantages}  
By introducing the unified size extension strategy, we significantly improve the alignment quality of depth maps when generating side RGB images, while maintaining computational efficiency. The following results were obtained under experimental conditions of $H=112$ and $W=112$.  
As observed in Fig.~\ref{figure_renderer}, prior to extension, there are significant size differences between images, causing obvious distortions and boundary errors in the RGB images.  
After the extension, all images have consistent sizes, leading to a significant improvement in the alignment of the generated RGB images, with boundary errors controlled to be less than $0.5$ pixels.  
Additionally, the memory and time overhead of this method is minimal, and the impact of the auxiliary images on the overall efficiency is negligible.  

This method, centered around unified depth map size, solves the alignment issue caused by size inconsistencies during the depth map generation process at the cost of introducing two auxiliary images. By combining depth map cropping and GPU parallel optimization, this strategy achieves a good balance between efficiency and effectiveness, providing a solid foundation for subsequent depth-based RGB image generation tasks. Fig.~\ref{renderer_ex} shows the performance of our renderer when rendering a hemisphere.

\section{Training Function for Face Reconstruction}
\begin{figure*}[h]
	\begin{center}
		\includegraphics[width=1.0\linewidth]{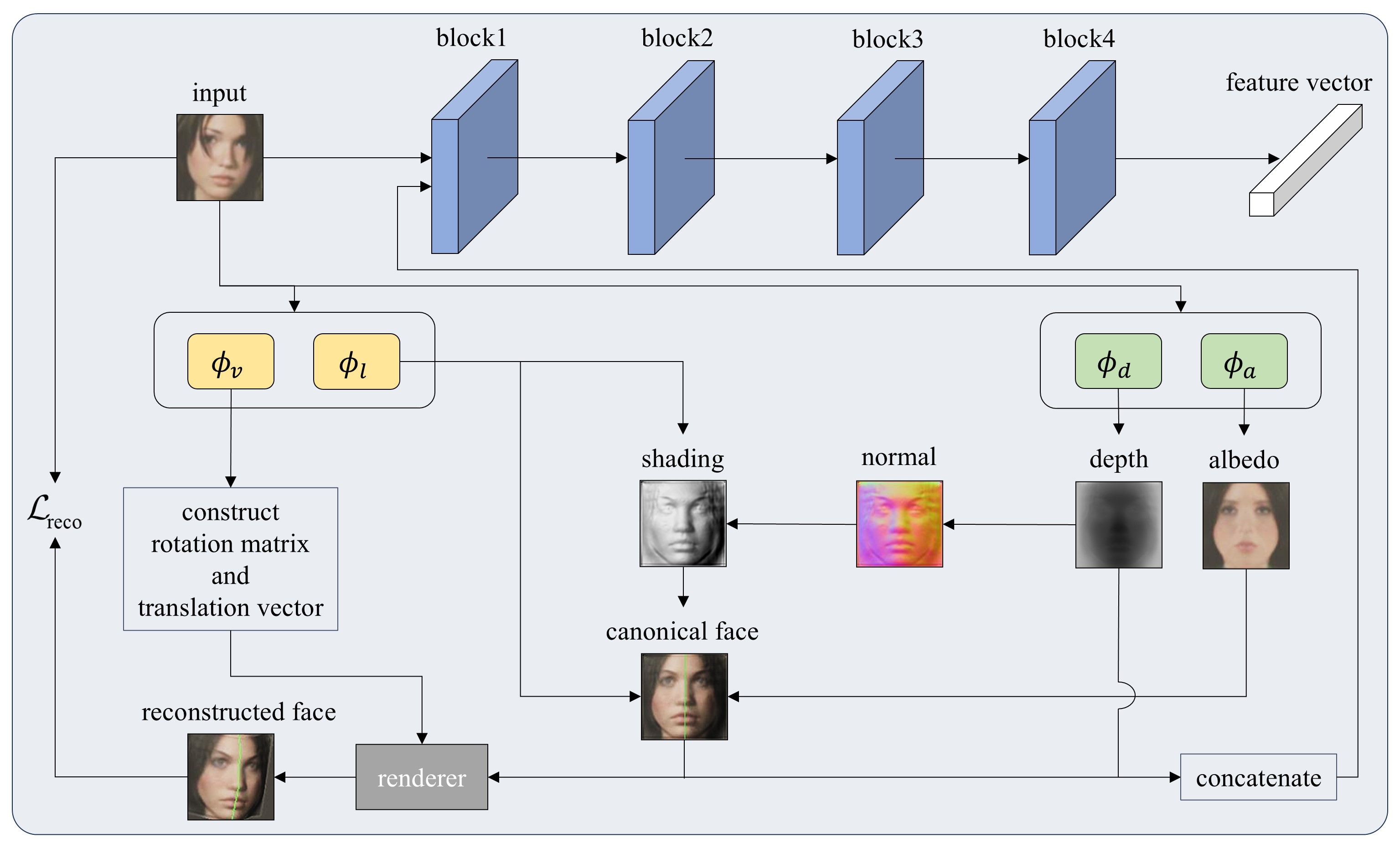}
	\end{center}
	\caption{The image shows the training process for face reconstruction and its connection to FR. To make the diagram less cluttered, the flowchart for the mirroring face reconstruction process has been omitted.}
	\label{figure_flowchart_reco}
\end{figure*}

In this study, a two-stage training strategy was adopted: First, a standard FR network was trained using the CelebA dataset, and then this structure was embedded into a ResNet-18 framework for further training by using a truncated version of MS1MV2. To optimize the joint task of FR and face reconstruction, a composite loss function with multiple objectives was designed to ensure the network's good performance across various aspects. Fig.~\ref{figure_flowchart_reco} illustrates the training flowchart for the integration of face reconstruction and FR. Below are the components included in the composite function.

\subsection{Face Recognition Loss ($\mathcal{L}_\text{FR}$)}
Since the core objective of the task is FR, the highest weight is assigned to the FR loss. This loss is computed based on feature vectors and ensures that the model maintains good recognition performance throughout the reconstruction process.

\subsection{Reconstruction Loss ($\mathcal{L}_\text{reco}$)}
To improve the accuracy of the reconstruction, the reconstruction loss ($\mathcal{L}_\text{reco}$) is introduced. This loss function quantifies the accuracy of image reconstruction and aims to minimize the difference between the reconstructed image and the target image, ensuring high visual quality and fidelity in the face reconstruction. $\mathcal{L}_\text{reco}$ consists of the following components: first, the loss between the reconstructed image and the original image is
\begin{equation}
	\mathcal{L}(\hat{\mathbf{I}}, \mathbf{I}, \sigma)=-\frac{1}{|\Omega|} \sum \ln \frac{1}{\sqrt{2} \sigma} \exp -\frac{\sqrt{2}|\hat{\mathbf{I}}-\mathbf{I}|}{\sigma}
\end{equation}
where $\Omega$ represents the valid region constrained by the mask obtained from the depth map. This is the negative log-likelihood of a factorized Laplace distribution applied to the reconstruction residuals. It is well-suited for face reconstruction tasks with large errors and noise. The horizontal flip of the depth map and albedo map's $\mathcal{L}(\hat{\mathbf{I}'}, \mathbf{I}, \sigma)$ is also computed. Next, the perceptual loss between the reconstructed image and the original image is
\begin{equation}
	\begin{split}
		&\mathcal{L}_{\mathrm{p}}^{(k)}\left(\hat{\mathbf{I}}, \mathbf{I}, \sigma^{(k)}\right)\\
		&=-\frac{1}{\left|\Omega_{k}\right|} \sum_{u v \in \Omega_{k}} \ln \frac{1}{\sqrt{2 \pi\left(\sigma^{(k)}\right)^{2}}} \exp -\frac{\left|e^{(k)}(\hat{\mathbf{I}})-e^{(k)}(\mathbf{I})\right|}{2\left(\sigma^{(k)}\right)^{2}}
	\end{split}
\end{equation}
For semantic difference, it follows a Gaussian distribution pattern. It also has a complete flip version.
During the calculation of the depth map, depth smoothness losses in both the vertical and horizontal directions are also computed.
\begin{equation}
	\begin{split}
		\mathcal{L}_{\text {smooth }}&=\frac{1}{N} \sum_{i=1}^{H-1} \sum_{j=1}^{W}\left|\frac{\mathbf{d}(i, j)-\mathbf{d}(i+1, j)}{\max \_ \text {depth }- \text { min\_depth }}\right|+\\
		&\frac{1}{N} \sum_{i=1}^{H} \sum_{j=1}^{W-1}\left|\frac{\mathbf{d}(i, j)-\mathbf{d}(i, j+1)}{\text { max\_depth }- \text { min\_depth }}\right|
	\end{split}
\end{equation}
The final reconstruction loss is
\begin{equation}
	\begin{split}
		&\mathcal{L}_\text{reco} = \mathcal{L}(\hat{\mathbf{I}}, \mathbf{I}, \sigma) + \lambda_\text{flip} \mathcal{L}(\hat{\mathbf{I}'}, \mathbf{I}, \sigma) +\\ &\lambda_\text{perc}\left(\mathcal{L}_{\mathrm{p}}^{(k)}\left(\hat{\mathbf{I}}, \mathbf{I}, \sigma^{(k)}\right) + \lambda_\text{flip} \mathcal{L}_{\mathrm{p}}^{(k)}\left(\hat{\mathbf{I}'}, \mathbf{I}, \sigma^{(k)}\right) \right)\\
		&+\lambda_\text{smooth}\mathcal{L}_{\text {smooth }}
	\end{split}
\end{equation}

\subsection{View Variance Loss ($\mathcal{L}_\text{view}$)}
In face reconstruction tasks, the model often tends to overly rely on the canonical image to minimize the reconstruction loss, thereby neglecting face features from different angles. To prevent this, an auxiliary loss function based on the variance of rotation angles was designed. By constraining the variance of the model's predictions across different viewpoints, this loss forces the model to maintain consistency in face features from multiple directions, improving its ability to handle face reconstruction from different perspectives.

Specifically, we divide the input 3D viewpoint information (i.e., rotation angles) into three directions (e.g., horizontal rotation, vertical rotation, and depth rotation) and calculate the variance for each direction. Then, these variances are penalized using the following loss function:
\begin{equation}
	\mathcal{L}_\text{view} = \text{ReLU}(v_1 - \text{var}_1) + \text{ReLU}(v_2 - \text{var}_2) + \text{ReLU}(v_3 - \text{var}_3)
\end{equation}
where $\text{var}_1$, $\text{var}_2$, and $\text{var}_3$ are the variances for the three rotation directions. By constraining the variances, we prevent the model from overly relying on a fixed viewpoint (e.g., the canonical view), ensuring that the model can effectively reconstruct face features from multiple perspectives, thus improving the diversity and robustness of the reconstruction. Among these, $\text{var}_2$ represents the variance of the rotation angle from the frontal view to the profile view, and it is the one with the largest range.

\subsection{Canonical Feature Loss ($\mathcal{L}_\text{FR}^\text{canon}$)}
This loss is computed by inputting the canonical image and calculating its feature vector, ensuring that the features of the canonical image can be effectively matched with the features of images from other viewpoints, thereby enhancing the model's representation capability for the canonical view. This loss helps to strengthen the alignment of the canonical view with the identity in the face reconstruction task, preventing the model from neglecting important features of the canonical image due to overtraining on the reconstruction task. The key difference between this loss and $\mathcal{L}_\text{FR}$ is that its input is the reconstructed canonical image.

\section{Supplementary Theory and Experiments on Proxy-Based Loss}  

\subsection{Formula for the Minimum Angle Between Uniformly Distributed Vectors in High-Dimensional Space}

In high-dimensional space, it is difficult to achieve a strictly uniform distribution. Nevertheless, a relatively uniform distribution similar to the minimum energy distribution in physics can be assumed. A uniform distribution increases inter-class differences, thus aiding classification. Therefore, in multi-class classification tasks, the model training process naturally tends to move towards a uniform distribution. Moreover, the closer the distribution of trained proxy vectors in high-dimensional space is to a uniform distribution, the greater the inter-class differences, leading to a better classification performance. Therefore, a uniform distribution can be treated as the expected outcome of training. A key characteristic of a uniform distribution is that the minimum angle between vectors remains constant.

\subsubsection{Uniform Distribution on a High-Dimensional Unit Sphere}  

Let $ \bm{u} $ and $ \bm{v} $ be two randomly distributed vectors on a $ d $-dimensional unit sphere. We focus on their inner product $ \bm{u} \cdot \bm{v} $, corresponding to their cosine similarity.  

\subsubsection{Concentration Phenomenon in High-Dimensional Space}  

In high-dimensional space, the cosine similarity between two random vectors, given by $ \cos \theta = \bm{u} \cdot \bm{v} $, approximately follows a normal distribution $ \mathcal{N}(0, \frac{1}{d}) $. It occurs because as the dimension $ d $ increases, the variance of the inner product approaches to $ \frac{1}{d} $. The proof of this result can be found in Section~\ref{prove1}.

\subsubsection{EVT Extreme Value Theory}  

To determine the minimum angle between the closest two vectors, the Extreme Value Theory (EVT) is utilized, which focuses on the minimum value among $ C $ random variables. Assume that $ X_1, X_2, \ldots, X_C $ are independent and identically distributed (i.i.d.) random variables, where $ X_i \sim \mathcal{N}(0, \frac{1}{d}) $.  

According to EVT, when $ C $ is large, the distribution of the minimum value can be approximated as:  
\begin{equation}
	\min_{i} X_i \approx \mu - \sigma \sqrt{2 \log C}
\end{equation}  
The detailed proof is provided in Section~\ref{prove2}. Here, $ \mu $ and $ \sigma $ represent the mean and standard deviation of the standard normal distribution. For a standard normal distribution $ \mathcal{N}(0, 1) $, $ \mu = 0 $ and $ \sigma = 1 $.

\subsubsection{Standardized Inner Product}  

Since $ X_i $ follows $ \mathcal{N}(0, \frac{1}{d}) $, it can be standardized as follows:
\begin{equation}
	\frac{X_i}{\sqrt{\frac{1}{d}}} \sim \mathcal{N}(0, 1)
\end{equation}  
Thus, the minimum value can be approximated as:  
\begin{equation}
	\min_{i} \frac{X_i}{\sqrt{\frac{1}{d}}} \approx \sqrt{\frac{1}{d}} \cdot (-\sqrt{2 \log C}) = -\sqrt{\frac{2 \log C}{d}}
\end{equation}  
Since we are concerned with the absolute value of the cosine similarity, the estimated minimum cosine similarity is given by:  
\begin{equation}
	\cos \theta_{\text{min}} \approx \sqrt{\frac{2 \log C}{d}}
\end{equation}  

\subsubsection{Explanation}  

This formula indicates that in high-dimensional space, the cosine similarity of the minimum angle among $ C $ uniformly distributed vectors is approximately $ \sqrt{\frac{2 \log C}{d}} $. This implies that the minimum angle is inversely proportional to the square root of $ d $ and directly proportional to the square root of $ \log C $.  

\subsection{Mean, Variance, and Quantile Function of Inner Products of Uniformly Distributed Vectors in High-Dimensional Space}  
\label{prove1}  

Let $ \bm{u} $ and $ \bm{v} $ be two randomly distributed vectors on a $ d $-dimensional unit sphere. We focus on their inner product $ \bm{u} \cdot \bm{v} $, given by:  
\begin{equation}
	\bm{u} \cdot \bm{v} = \sum_{i=1}^d u_i v_i
\end{equation}  

\subsubsection{Mean}  

Since $ \bm{u} $ and $ \bm{v} $ are uniformly distributed, their components $ u_i $ and $ v_i $ are independent and have a mean of 0. Therefore, the following holds:
\begin{equation}
	\mathbb{E}[u_i v_i] = \mathbb{E}[u_i] \mathbb{E}[v_i] = 0
\end{equation}  
leading to:  
\begin{equation}
	\mathbb{E}[\bm{u} \cdot \bm{v}]=\mathbb{E}\left[\sum_{i=1}^{d} u_{i} v_{i}\right]=\sum_{i=1}^{d} \mathbb{E}\left[u_{i} v_{i}\right]=\sum_{i=1}^{d} 0=0
\end{equation}  
Thus, the expected value of the inner product $ \bm{u} \cdot \bm{v} $ is 0.  

\subsubsection{Variance}  

To compute the variance $ \text{Var}(\bm{u} \cdot \bm{v}) $, the expectation $ \mathbb{E}[(\bm{u} \cdot \bm{v})^2] $ must first be calculated.
\begin{equation}
	(\bm{u} \cdot \bm{v})^2 = \left( \sum_{i=1}^d u_i v_i \right)^2 = \sum_{i=1}^d u_i^2 v_i^2 + 2 \sum_{i < j} u_i v_i u_j v_j
\end{equation}  
Since $ u_i $ and $ v_i $ are independent and uniformly distributed variables, and given that each variable has an expected value of 0, we can determine that the variance of $ u_i $ and $ v_i $ is $ \frac{1}{d} $ based on the unit vector definition. From the expectation property, the following is obtained:  
\begin{equation}
	\mathbb{E}[u_i^2] = \mathbb{E}[v_i^2] = \frac{1}{d}
\end{equation}  
Since $ u_i $ and $ v_i $ are independent, it follows that:  
\begin{equation}
	\mathbb{E}[u_i^2 v_i^2] = \mathbb{E}[u_i^2] \mathbb{E}[v_i^2] = \left( \frac{1}{d} \right)^2
\end{equation}  
For the cross terms $ \sum_{i < j} u_i v_i u_j v_j $, the following holds:
\begin{equation}
	\mathbb{E}[u_i v_i u_j v_j] = \mathbb{E}[u_i v_i] \mathbb{E}[u_j v_j] = 0
\end{equation}  
because $ u_i $ and $ v_i $ are independent variables with a mean of 0.  

Thus, the following is obtained:
\begin{equation}
	\mathbb{E}[(\bm{u} \cdot \bm{v})^2] = \sum_{i=1}^d \mathbb{E}[u_i^2 v_i^2] = d \left( \frac{1}{d} \right)^2 = \frac{1}{d}
\end{equation}  
Therefore, the variance of the inner product is given by:  
\begin{equation}
	\text{Var}(\bm{u} \cdot \bm{v}) = \mathbb{E}[(\bm{u} \cdot \bm{v})^2] - (\mathbb{E}[\bm{u} \cdot \bm{v}])^2 = \frac{1}{d}
\end{equation}  

\subsubsection{Quantile Function}  

The quantile function is the inverse of the cumulative distribution function (CDF). For a random variable $ X $, its cumulative distribution function $ F(x) $ is defined as $ F(x) = P(X \leq x) $. The quantile function $ Q(p) $ is the value $ x $ that satisfies $ P(X \leq Q(p)) = p $.  

For the standard normal distribution $ \mathcal{N}(0, 1) $, the quantile function $ Q(p) $ gives the value where the cumulative distribution function equals $ p $. For example, the 50\% quantile of the standard normal distribution is 0 because there is a 50\% probability that $ X $ is less than or equal to 0.  

\subsection{Derivation of the Approximation $ \sqrt{2 \log C} $}  
\label{prove2}  

The derivation of $ \sqrt{2 \log C} $ originates from extreme value theory in high-dimensional space. The detailed derivation process is as follows:  

\subsubsection{Distribution of Cosine Similarity}  

First, consider $ N $ uniformly distributed unit vectors in a $ d $-dimensional space. For any two unit vectors $ \bm{u} $ and $ \bm{v} $, their inner product $ \bm{u} \cdot \bm{v} $ follows a normal distribution $ \mathcal{N}(0, \frac{1}{d}) $.  

\subsubsection{Standardized Cosine Similarity}  

To facilitate processing, the inner product is standardized:
\begin{equation}
	Z = d \cdot (\bm{u} \cdot \bm{v})
\end{equation}  
Then, $ Z $ follows a standard normal distribution $ \mathcal{N}(0, 1) $.  

\subsubsection{Distribution of the Minimum Value}  

Consider $ N $ random variables $ Z_1, Z_2, \ldots, Z_C $, each following a standard normal distribution $ \mathcal{N}(0, 1) $. We aim to determine the minimum value, given by $ \min(Z_1, Z_2, \ldots, Z_C) $.  

According to extreme value theory, when $ C $ is large, the distributions of both the maximum and minimum values can be approximated as:  
\begin{equation}
	\min(Z_1, Z_2, \ldots, Z_C) \approx -\sqrt{2 \log C}
\end{equation}  
This result can be derived as follows:  

\paragraph{Quantile Function of the Standard Normal Distribution}  

The quantile function $ Q(p) $ of a standard normal distribution is defined as the function satisfying $ P(Z \leq Q(p)) = p $. For $ Z \sim \mathcal{N}(0, 1) $, the quantile function $ Q(p) $ can be approximated as:  
\begin{equation}
	Q(p) \approx \sqrt{2 \log \left(\frac{1}{1 - p}\right)}
\end{equation}  
The proof of this result can be found in Section~\ref{prove3}.  

\paragraph{Quantile of the Minimum Value}  

For $ N $ independent and identically distributed (i.i.d.) standard normal random variables $ Z_1, Z_2, \ldots, Z_C $, the CDF of the minimum value, $ F_{\min}(x) $, is given by:  
\begin{equation}
	F_{\min}(x) = 1 - (1 - F(x))^C
\end{equation}  
where $ F(x) $ is the CDF of the standard normal distribution. The quantile function of the minimum value, $ Q_{\min}(p) $, is given by:  
\begin{equation}
	Q_{\min}(p) = Q\left(1 - \frac{p}{C}\right)
\end{equation}  
The proof of these steps can be found in Section~\ref{prove6}. Using an approximation for the quantile function:  
\begin{equation}
	Q_{\min}(p)  \approx \sqrt{2 \log \left(\frac{C}{p}\right)}
\end{equation}  
As $ p \to 1 $, it follows that:
\begin{equation}
	Q_{\min}(1) \approx -\sqrt{2 \log C}
\end{equation}  
This approximation indicates that the minimum value among $ C $ standard normal random variables is approximately around $ -\sqrt{2 \log C} $.  

\subsection{Derivation of $ Q(p) \approx \sqrt{2 \log \left(\frac{1}{1 - p}\right)} $}  
\label{prove3}  

The approximation $ Q(p) \approx \sqrt{2 \log \left(\frac{1}{1 - p}\right)} $ is valid when the probability $ p $ is very close to 1. By analyzing the properties of the cumulative distribution function and the quantile function, it derives that approximation arises from the asymptotic behavior in the tail of the distribution.

\subsubsection{Quantile Function $ Q(p) $ of the Standard Normal Distribution}  

For a standard normal distribution $ Z \sim \mathcal{N}(0, 1) $, the CDF $ F(z) $ is given by:  
\begin{equation}
	F(z) = \frac{1}{2} \left[ 1 + \text{erf} \left( \frac{z}{\sqrt{2}} \right) \right]
\end{equation}  
where $ \text{erf}(x) $ is the error function. The proof of this result can be found in Section~\ref{prove4}.  

The quantile function $ Q(p) $ is the inverse function of $ F(z) $, i.e.,  
\begin{equation}
	Q(p) = F^{-1}(p)
\end{equation}  
When $ p $ is very close to 1 (e.g., $ p \to 1 $), we seek an approximation to describe the behavior of $ Q(p) $.  

\subsubsection{Approximate Derivation}  

\paragraph{Tail Behavior of the Cumulative Distribution Function}  

In the tail region of standard normal distribution (i.e., when $ z $ is large), we have the following asymptotic approximation:  
\begin{equation}
	1 - F(z) \approx \frac{1}{\sqrt{2\pi}} \frac{1}{z} e^{-z^2/2}
\end{equation}  
This approximation arises from the asymptotic properties of the normal distribution tail.  

\paragraph{Finding the Inverse Function}  

We are interested in the behavior of the quantile function $ Q(p) $ as $ p $ approaches 1.  

Let $ p = 1 - \epsilon $, where $ \epsilon $ is small, i.e., $ \epsilon \to 0 $. We seek the value of $ z $ that satisfies $ F(z) = p $, corresponding to $ Q(p) $.  
\begin{equation}
	\epsilon = 1 - p = 1 - F(z) \approx \frac{1}{\sqrt{2\pi}} \frac{1}{z} e^{-z^2/2}
\end{equation}  
Taking the logarithm:  
\begin{equation}
	\log \epsilon \approx \log \left( \frac{1}{\sqrt{2\pi}} \frac{1}{z} \right) - \frac{z^2}{2}
\end{equation}  
Since $ \epsilon $ is very small, $ \log \epsilon $ is a large negative number. Ignore the constant terms in the logarithm:  
\begin{equation}
	\log \epsilon \approx - \frac{z^2}{2} - \log z - \frac{1}{2} \log 2\pi
\end{equation}  
For large $ z $, the term $ \log z $ is negligible compared to $ z^2/2 $, then we approximate:  
\begin{equation}
	\log \epsilon \approx - \frac{z^2}{2}
\end{equation}  
Thus,  
\begin{equation}
	z^2 \approx -2 \log \epsilon
\end{equation}  
Returning to $ p = 1 - \epsilon $:  
\begin{equation}
	z^2 \approx -2 \log (1 - p)
\end{equation}  
leading to:  
\begin{equation}
	z \approx \sqrt{-2 \log (1 - p)} \approx \sqrt{2 \log \left( \frac{1}{1 - p} \right)}
\end{equation}  

\paragraph{Deriving the Approximate Formula}  

Thus, when $ p $ nearly close to 1, the quantile function $ Q(p) $ of the standard normal distribution can be approximated as:  
\begin{equation}
	Q(p) \approx \sqrt{2 \log \left( \frac{1}{1 - p} \right)}
\end{equation}  
This approximation explains how the quantile function value grows rapidly as probability $ p $ approaches 1. This rapid growth is significant in many applications, such as in extreme value theory.  

\subsection{Asymptotic Properties of the Normal Distribution Tail and the Error Function}  
\label{prove4}  

\subsubsection{Asymptotic Properties of the Normal Distribution Tail}  

The asymptotic properties of the normal distribution tail refer to the behavior of the cumulative distribution function $ F(z) $ and the PDF $ f(z) $ when $ z $ is large (i.e., in the tail region of the normal distribution). Specifically, for a standard normal distribution $ \mathcal{N}(0,1) $, when $ z $ is very large (right tail) or very small (left tail), the distribution function can be approximated by using an asymptotic expansion.  

The PDF $ f(z) $ of the standard normal distribution is given by:  
\begin{equation}
	f(z) = \frac{1}{\sqrt{2\pi}} e^{-z^2/2}
\end{equation}  
The cumulative distribution function $ F(z) $ is given by:  
\begin{equation}
	F(z) = \int_{-\infty}^z f(t) \, dt
\end{equation}  
In the tail region (for large $ z $), the cumulative distribution function $ F(z) $ can be approximated using the tail behavior of the density function:  
\begin{equation}
	1 - F(z) = P(Z > z)
\end{equation}  
Using an asymptotic expansion, the following is obtained:
\begin{equation}
	1 - F(z) \approx \frac{1}{\sqrt{2\pi}} \frac{1}{z} e^{-z^2/2}
\end{equation}  
This approximation originates from the tail behavior of the normal density function $ f(z) $. The detailed proof can be found in Section~\ref{prove5}.  

\subsubsection{Error Function}  

The error function $ \text{erf}(x) $ is a special function widely used in probability theory and statistics. It is defined as:  
\begin{equation}
	\text{erf}(x) = \frac{2}{\sqrt{\pi}} \int_0^x e^{-t^2} \, dt
\end{equation}  

Properties of the error function:  
\begin{equation}
	\begin{split}
		&\left(1\right) \ \text{erf}(0) = 0\\
		&\left(2\right) \ \text{erf}(\infty) = 1\\
		&\left(3\right) \ \text{erf}(-x) = -\text{erf}(x)
	\end{split}
\end{equation}  

For the standard normal distribution, its cumulative distribution function $ F(z) $ can be expressed in terms of the error function as:  
\begin{equation}
	F(z) = \frac{1}{2} \left[ 1 + \text{erf} \left( \frac{z}{\sqrt{2}} \right) \right]
\end{equation}  

\subsection{Proof of $ 1-F(z) \approx \frac{1}{\sqrt{2 \pi}} \frac{1}{z} e^{-z^{2} / 2} $}  
\label{prove5}  

For a standard normal distribution $ Z \sim \mathcal{N}(0, 1) $, its PDF $ f(z) $ is given by:  
\begin{equation}
	f(z) = \frac{1}{\sqrt{2\pi}} e^{-z^2/2}
\end{equation}  
The cumulative distribution function $ F(z) $ is:  
\begin{equation}
	F(z) = \int_{-\infty}^z f(t) \, dt
\end{equation}  
We are interested in the tail behavior of the cumulative distribution function, i.e., $ 1 - F(z) $ as $ z \to \infty $:  
\begin{equation}
	1 - F(z) = P(Z > z) = \int_z^\infty f(t) \, dt
\end{equation}  
Substituting $ f(t) = e^{-t^2/2} $, the following is obtained:
\begin{equation}
	1 - F(z) = \frac{1}{\sqrt{2\pi}} \int_z^\infty e^{-t^2/2} \, dt
\end{equation}  

We use the method of integration by parts:  
Let $ u = 1 $ and $ dv = \frac{1}{\sqrt{2\pi}} e^{-t^2 / 2} \, dt $, then $ du = 0 $ and $ v = \frac{1}{t} e^{-t^2 / 2} $. Applying the integration by parts formula $ \int u \, dv = uv - \int v \, du $, we get:  
\begin{equation}
	1 - F(z) = \left. \frac{1}{t} \frac{1}{\sqrt{2\pi}} e^{-t^2 / 2} \right|_z^\infty = \frac{1}{z} \frac{1}{\sqrt{2\pi}} e^{-z^2 / 2}
\end{equation}  
Thus, the final result is obtained:
\begin{equation}
	1 - F(z) = \frac{1}{\sqrt{2 \pi}} \frac{1}{z} e^{-z^2 / 2}
\end{equation}  

\subsection{Derivation of the Distribution Function and Quantile Function of the Minimum of Random Variables}  
\label{prove6}  

Suppose $ Z_1, Z_2, \ldots, Z_C $ are $ C $ independent and identically distributed (i.i.d.) standard normal random variables.  

\subsubsection{Derivation of the Distribution Function $ F_{\min}(x) $}  

The distribution function $ F_{\min}(x) $ of the minimum value represents the probability that the minimum is less than or equal to $ x $. In other words:  
\begin{equation}
	F_{\min}(x) = P(\min(Z_1, Z_2, \ldots, Z_C) \leq x)
\end{equation}  
This is equivalent to the probability of the complement event where all $ Z_i $ are greater than $ x $:  
\begin{equation}
	F_{\min}(x) = 1 - P(Z_1 > x, Z_2 > x, \ldots, Z_C > x)
\end{equation}  
Since the $ Z_i $ are independent and identically distributed, the following can be written: 
\begin{equation}
	\begin{split}
		&P(Z_1 > x, Z_2 > x, \ldots, Z_C > x) \\ 
		&= P(Z > x)^C = (1 - F(x))^C
	\end{split}
\end{equation}  
where $ F(x) $ is the CDF of the standard normal distribution. Therefore:  
\begin{equation}
	F_{\min}(x) = 1 - (1 - F(x))^C
\end{equation}  

\subsubsection{Derivation of the Quantile Function $ Q_{\min}(p) $}  

The quantile function $ Q_{\min}(p) $ is the value that satisfies $ F_{\min}(Q_{\min}(p)) = p $. That is,  
\begin{equation}
	F_{\min}(Q_{\min}(p)) = p
\end{equation}  
Substituting the expression for $ F_{\min}(x) $:  
\begin{equation}
	1 - (1 - F(Q_{\min}(p)))^C = p
\end{equation}  
Solving for $ Q_{\min}(p) $:  
\begin{equation}
	\begin{split}
		&(1 - F(Q_{\min}(p)))^C = 1 - p\\
		&1 - F(Q_{\min}(p)) = (1 - p)^{1/C}\\
		&F(Q_{\min}(p)) = 1 - (1 - p)^{1/C}
	\end{split}
\end{equation}  
Since $ Q(p) $ is the quantile function of the standard normal distribution, satisfying $ F(Q(p)) = p $, the following is obtained:
\begin{equation}
	Q_{\min}(p) = Q(1 - (1 - p)^{1/C})
\end{equation}  
For $ p $ close to 1, $ 1 - p $ is very small, allowing the approximation:  
\begin{equation}
	Q_{\min}(p) \approx Q\left(1 - \frac{p}{C}\right)
\end{equation}  
This follows from the approximation $ 1 - (1 - p)^{1/C} \approx \frac{p}{C} $ when $ p $ is close to 1.  

Thus, the distribution function and the quantile function of the minimum value can be expressed as:  
\begin{equation}
	\begin{split}
		&F_{\min}(x) = 1 - (1 - F(x))^C\\
		&Q_{\min}(p) = Q\left(1 - \frac{p}{C}\right)
	\end{split}
\end{equation}  

\subsection{Numerical Computation}  

\subsubsection{Application of the Formula}  

For subMS1MV2, $ C = 70722 $. These vectors, uniformly distributed in a 512-dimensional space, can be used to estimate the minimum angle using the following computation:  
\begin{equation}
	\cos \theta_{\text{min}} \approx \sqrt{\frac{2 \log C}{d}}
\end{equation}  
Substituting the values into the formula:  
\begin{equation}
	\cos \theta_{\text{min}} \approx \sqrt{\frac{2 \times 11.16}{512}} \approx \sqrt{0.0436} \approx 0.2089
\end{equation}  
Then, calculating the corresponding angle:  
\begin{equation}
	\theta_{\text{min}} \approx 1.36 \text{ rad}
\end{equation}  
Converting radians to degrees:  
\begin{equation}
	\theta_{\text{min}} \approx 77.94^\circ
\end{equation}  
Thus, for 70722 vectors uniformly distributed in a 512-dimensional space, the minimum angle between the closest two vectors is approximately $ 77.94^\circ $.  

\begin{figure}[h]
	\begin{center}
		\includegraphics[width=1.0\linewidth]{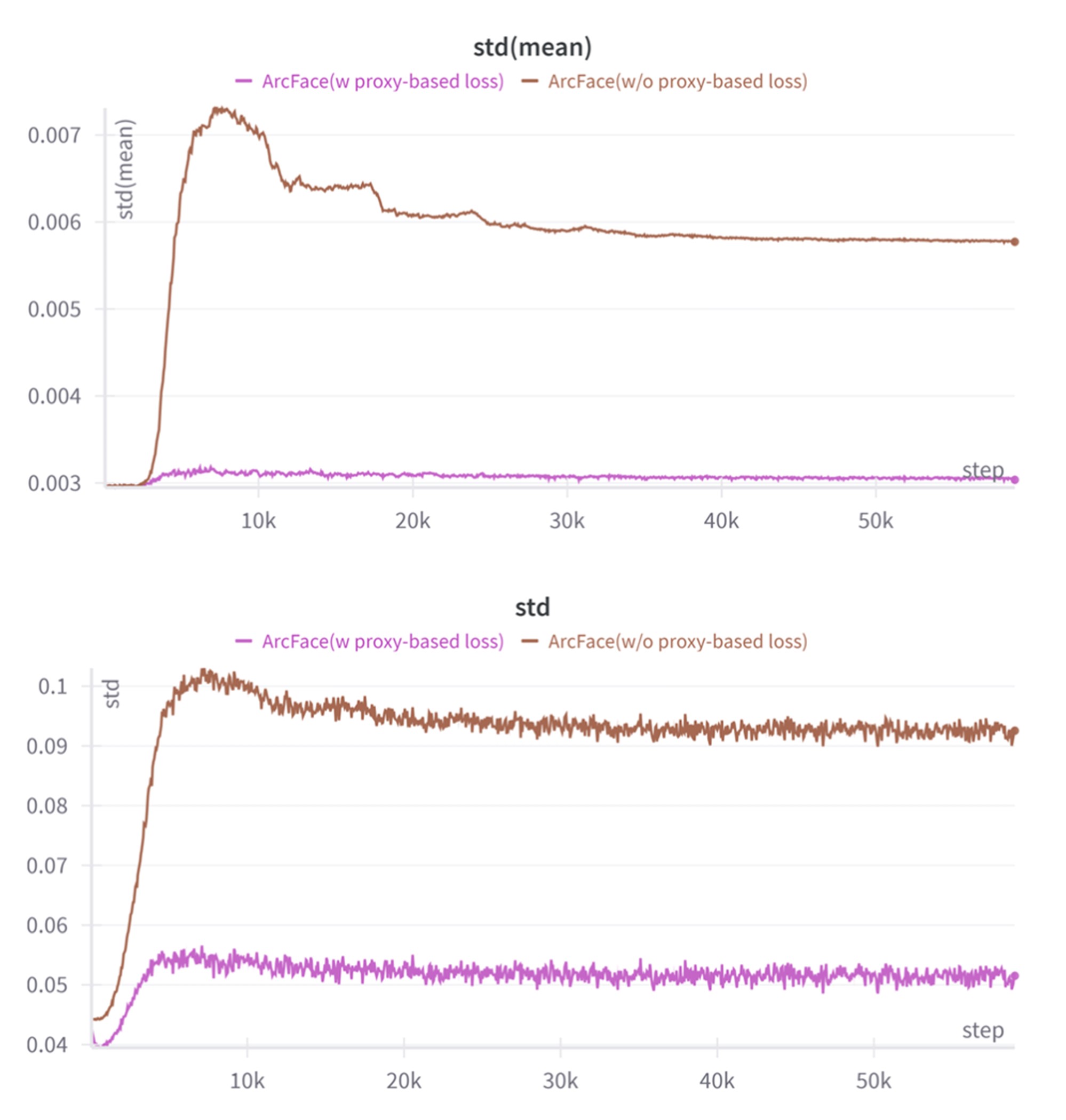}
	\end{center}
	\caption{The image of ArcFace's std and std(mean) changing with steps before and after adding the proxy-based loss.}
	\label{std}
\end{figure}

\subsubsection{Variance and Standard Deviation}  

For two random vectors $ \bm{u} $ and $ \bm{v} $ uniformly distributed on a $ d $-dimensional unit sphere, their inner product $ \bm{u} \cdot \bm{v} $ approximately follows a normal distribution $ \mathcal{N}(0, 1/d) $. Therefore, the variance of the cosine similarity is given by $ 1/d $. Specifically,  
\begin{equation}
	\text{Var}(\cos \theta) = \frac{1}{d}
\end{equation}  
The standard deviation is the square root of the variance:  
\begin{equation}
	\text{Std}(\cos \theta) = \sqrt{\text{Var}(\cos \theta)} = \sqrt{\frac{1}{d}}
	\label{eq_std}
\end{equation}  
Applying Eq.~\ref{eq_std} to the specific case where $ d = 512 $, the variance is:  
\begin{equation}
	\text{Var}(\cos \theta) = \frac{1}{512} \approx 0.001953
\end{equation}  
And the standard deviation is:  
\begin{equation}
	\text{Std}(\cos \theta) = \sqrt{\frac{1}{512}} \approx 0.0442
\end{equation}  

\subsubsection{Computing the Cosine Values of $ \theta_{\text{min}}/2 $ and $ \theta_{\text{min}}/4 $}  

\begin{equation}
	\theta_{\text{min}}/2 \approx \frac{1.358}{2} = 0.679 \text{ rad}
\end{equation}  
The cosine value is $ \cos(0.679) $.  

\begin{equation}
	\theta_{\text{min}}/4 \approx \frac{1.358}{4} = 0.3395 \text{ rad}
\end{equation}  
The cosine value is $ \cos(0.3395) $.  

Now, computing these cosine values:  
\begin{equation}
	\begin{split}
		&\cos(0.68) \approx 0.78\\
		&\cos(0.34) \approx 0.94
	\end{split}
\end{equation}  
Thus, the final results are:  
\begin{equation}
	\begin{split}
		&\cos(\theta_{\text{min}}/2) \approx 0.78\\
		&\cos(\theta_{\text{min}}/4) \approx 0.94
	\end{split}
\end{equation}  

It implies that when the cosine similarity between a proxy and its positive sample exceeds about $ 0.78 $, it can be considered as a fully successful classification.

\subsection{Experiments on Proxy-Based Loss}  

\begin{figure*}
	\begin{center}
		\includegraphics[width=1.0\linewidth]{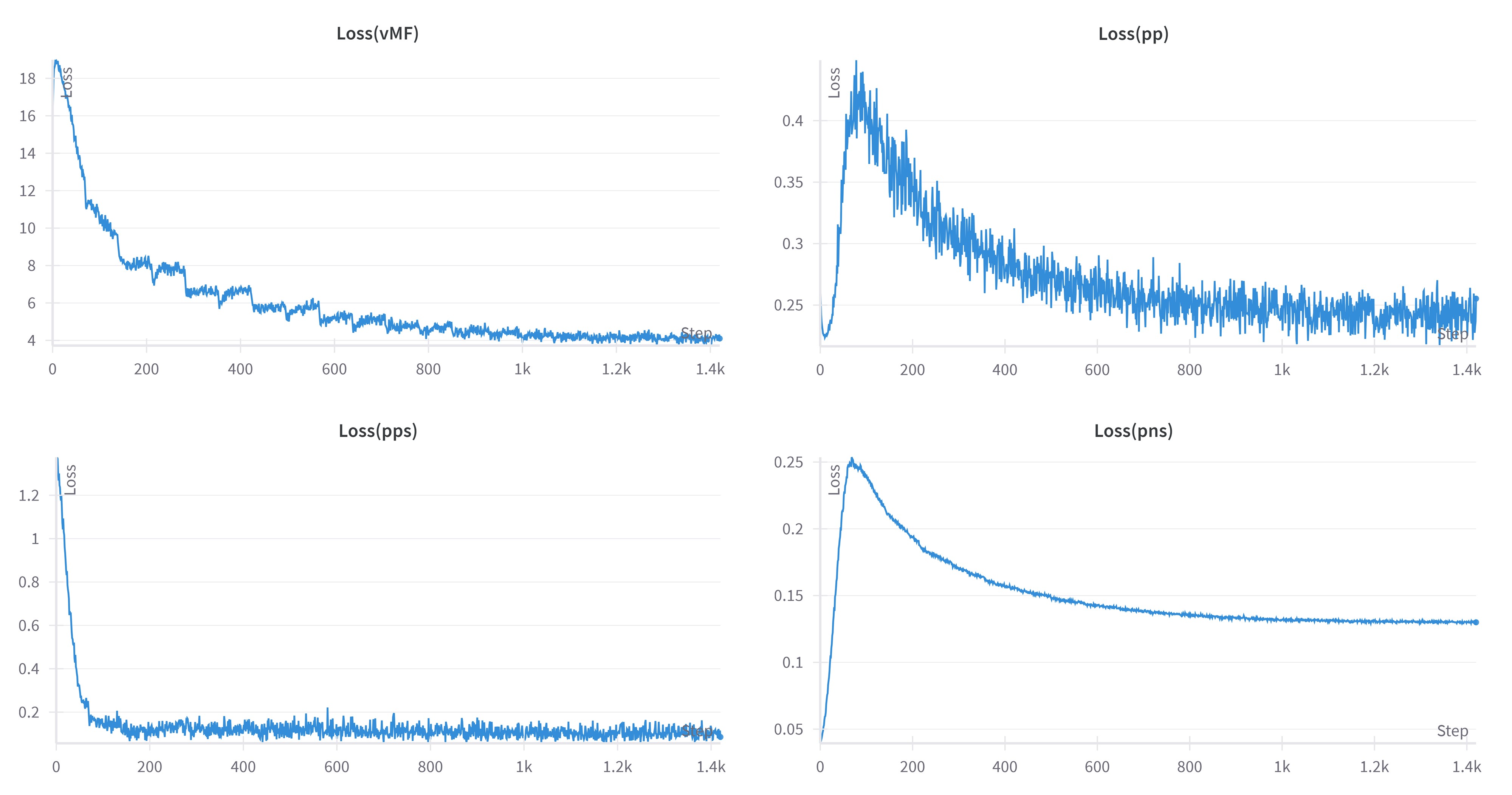}
	\end{center}
	\caption{This picture shows how the value of each component of $\mathcal{L}_\text{FR}$ changes with steps during a complete training process.}
	\label{loss}
\end{figure*}

We tracked the proxy feature vectors to observe whether the values of $ \cos \left(W_{y_i}, W_j\right) $ align with the inferred results. The calculations were performed by using the following formulas, specifically for $ \text{std}_\text{mean} $ and $ \text{std} $. 
\begin{small}
	\begin{equation}
		\begin{split}
			&\text{std}_\text{mean} = \\ 
			&\sqrt{\frac{\sum_{i=1}^{\left(N+N_{\text {unique }}\right)\left(N+N_{\text {unique }}-1\right) / 2} \operatorname{relu}^2\left(\cos\left(\mathbf{W}_{y_i}, \mathbf{W}_j\right)-\sqrt{\frac{2 \log C}{d}}\right)}{\left(N+N_{\text {unique }}\right)\left(N+N_{\text {unique }}-1\right) / 2}}
		\end{split}
	\end{equation}
\end{small}
\begin{equation}
	\begin{small}
		\text{std} = \sqrt{\frac{\sum_{i=1}^{\left(N+N_{\text {unique }}\right)\left(N+N_{\text {unique }}-1\right) / 2} \cos^2 \left(\mathbf{W}_{y_i}, \mathbf{W}_j\right)}{\left(N+N_{\text {unique }}\right)\left(N+N_{\text {unique }}-1\right) / 2}}
	\end{small}
\end{equation}

The formula for std without regard of the mean, assume it to be zero. We tracked these two variables as they changed with steps before and after incorporating the proxy-based loss in Fig.~\ref{std}. Observing the results, it is evident that after adding the proxy-based loss, ArcFace’s proxy feature vectors are significantly closer to uniform distributed.  
No matter how much it converges, the std cannot be lower than $ \sqrt{\frac{1}{d}} \approx 0.0442 $. Similarly, the nearest neighbor mean also approaches $ \sqrt{\frac{2 \log C}{d}} $.  
Therefore, we can conclude that the inclusion of proxy-based loss makes the proxy feature vectors more uniformly distributed.  

Fig.~\ref{loss} shows the variation of $\mathcal{L}_\text{vMF}$, $\mathcal{L}_\text{pps}$, $\mathcal{L}_\text{pp}$, and $\mathcal{L}_\text{pns}$ with respect to the training step when $\lambda_\text{pps} = 5$, $\lambda_\text{pp} = 150$, and $\lambda_\text{pns} = 20$. From the results, it is easy to observe that $\lambda_\text{pns}$ is the only loss function that exhibits relatively stable values. Initially, due to random distribution, its value is close to $0$. After training, its value increases rapidly because the proxy vectors begin to be associated with individual identities. At this point, its variation trend becomes similar to that of $\lambda_\text{pp}$, both increasing sharply and then gradually decrease, approaching to a uniform distribution, but never quite achieving it. This is a limited relationship: if the network is expanded, this limiting value will diminish. However, no matter how deep the network is or how large the dataset is, this limiting value will always greater than $0$. The value of $\lambda_\text{pps}$ is initially very large because $\left.\cos \left(W_{y_i}, z_i\right)_{\text {mid }}\right|_{e-1}$ has a minimum value of $0.5$. This minimum value causes the large loss of the randomly distributed feature vectors at the beginning.

\section{Sample-based Loss}

\begin{table}
	\caption{The ablation study of ArcFace on $\mathcal{L}_\text{sns}$.}
	\begin{center}
		\begin{tabular}{|l|c|ccc|}
			\hline
			\multicolumn{1}{|c|}{Method} & \multicolumn{1}{c|}{$\mathcal{L}_\text{nps}$} & High-quality & IJB-B/C & TinyFace \\ 
			\hline
			ArcFace~\cite {deng2019arcface} & $w/o$ & $\bm{93.05}$ & $57.03$ &$\bm{68.05}$\\
			($m=0.5$) & $w/$ & $92.66$ & $\bm{57.08}$ & $67.89$\\
			\hline
		\end{tabular}
		\\
	\end{center}
	\label{table:ss}
\end{table}

In fact, since we have considered pps, pns and pp, it is natural for us to study the possibility of ss, a loss function within the samples themselves as well. Experiments were conducted and the final results indicated that this approach may not effective. The proposed loss function is defined as:  
\begin{equation}
	\mathcal{L}_\text{sns} = \lambda_\text{sns} \frac{\sum_{i=1}^{N_{\text {unique}}\left(N_{\text {unique }}-1\right) / 2} \cos \left(\bm{z}_i, \bm{z}_j\right)}{N_{\text {unique }}\left(N_{\text {unique }}-1\right) / 2}
\end{equation}  
where "sns" stands for Sample and Negative Sample, and it applies when $ y_i \neq y_j $. $ \lambda_\text{sns} = 150 $ is set.

\begin{figure}[h]
	\begin{center}
		\includegraphics[width=1.0\linewidth]{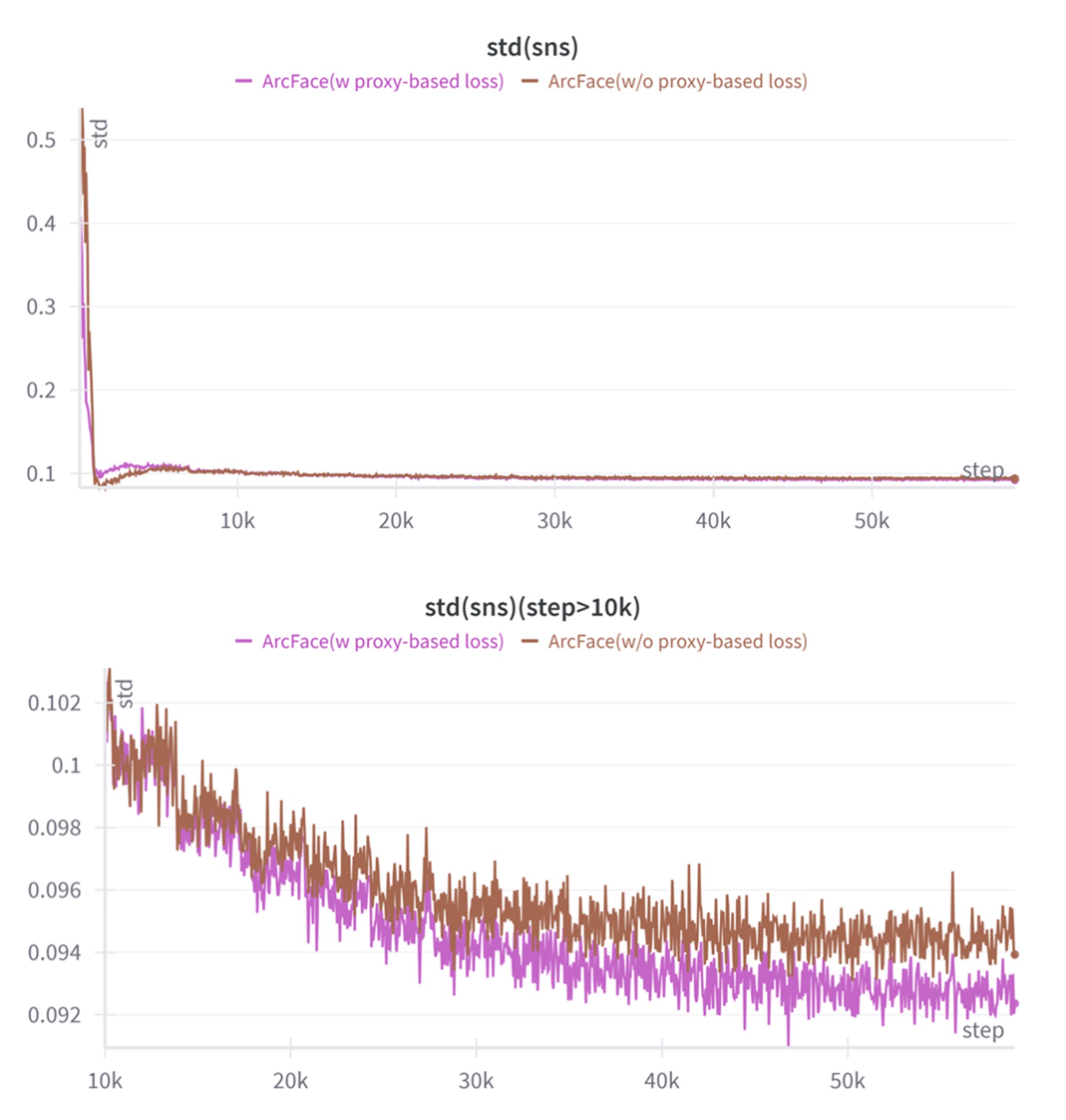}
	\end{center}
	\caption{The image of the variation of ArcFace's $\text{std}_\text{sns}$ with respect to step.}
	\label{sns}
\end{figure}

From the results in Table~\ref{table:ss}, it is evident that $ \mathcal{L}_\text{sns} $ does not improve performance, making us abandon this loss function. Although this loss function is not directly applicable, the evolution of the standard deviation of the cosine similarity between samples and negative samples with training steps can still be observed. The calculation method is as follows:  
\begin{equation}
	\begin{split}
		\text{std}_\text{sns} = 
		\sqrt{\frac{\sum_{i=1}^{N_{\text {unique}}\left(N_{\text {unique }}-1\right) / 2} \cos^2 \left(\bm{z}_i, \bm{z}_j\right)}{N_{\text{unique}}\left(N_{\text{unique}}-1\right) / 2}}
	\end{split}
\end{equation}  
In Fig.~\ref{sns}, the upper and lower plots are based on the same data source, with the second plot restricting the x-axis to $ \text{step} > 10k $. Observing the second plot, it can be found that proxy-based loss also positively impacts sample-based loss, even though no direct penalty was applied to the sample-based loss.

\section{Experiments on Proxy-based Loss Parameters}

We conducted comparative experiments on the parameter values of $\mathcal{L}_\text{pps}$, $\mathcal{L}_\text{pp}$, and $\mathcal{L}_\text{pns}$. It is clear that the experimental results are significantly better when $\mathcal{L}_\text{pps}=5$, $\mathcal{L}_\text{pp}=150$, and $\mathcal{L}_\text{pns}=20$ compared to other parameter values in Table~\ref{table:lambda}. After careful inspection, it shows that even when the parameter values are not optimal, the performance on IJB-B/C still outperforms ArcFace without the proxy-based loss.

Moreover, $\mathcal{L}_\text{pps}$ clearly has its own characteristics and functions. As shown in Table~\ref{table:lambda_uamf}, the larger the weight of $\mathcal{L}_\text{pps}$, the more significant the improvement in recognizing high-quality normal samples. However, the effect on recognizing hard high-quality samples and low-quality samples becomes less noticeable.

\begin{table}[h]
	\caption{Comparison of ablation experiments on elements $\mathcal{L}_\text{pps}$, $\mathcal{L}_\text{pp}$, and $\mathcal{L}_\text{pns}$ in the ArcFace~\cite {deng2019arcface}($m=0.5$).}
	\begin{center}
		\begin{tabular}{|c|c|c|ccc|}
			\hline
			\multicolumn{1}{|c|}{$\lambda_\text{pps}$} & \multicolumn{1}{c|}{$\lambda_\text{pp}$} & \multicolumn{1}{c|}{$\lambda_\text{pns}$} & High-quality & IJB-B/C & TinyFace \\ 
			\hline
			$0$ & $0$ & $0$ & $93.05$ & $57.03$ &$68.05$\\
			\hline
			$0$ & $150$ & $20$ & $93.04$ & $57.98$ &$67.52$\\
			$5$ & $150$ & $20$ & $93.16$ & $\bm{59.39}$ &$\bm{67.81}$\\
			$10$ & $150$ & $20$ & $\bm{93.24}$ & $58.49$ &$67.62$\\
			\hline
			$5$ & $100$ & $20$ & $93.15$ & $58.16$ &$\bm{68.35}$\\
			$5$ & $150$ & $20$ & $\bm{93.16}$ & $\bm{59.39}$ &$67.81$\\
			$5$ & $200$ & $20$ & $93.04$ & $59.02$ &$68.29$\\
			\hline
			$5$ & $150$ & $10$ & $93.02$ & $58.05$ &$67.27$\\
			$5$ & $150$ & $20$ & $93.16$ & $\bm{59.39}$ &$67.81$\\
			$5$ & $150$ & $30$ & $\bm{93.17}$ & $58.70$ &$\bm{67.95}$\\
			\hline
		\end{tabular}
		\\
	\end{center}
	\label{table:lambda}
\end{table}

\begin{table}[h]
	\caption{Comparison of ablation experiments on elements $\mathcal{L}_\text{pps}$ in the LH$^2$Face(UAMF + RECO + $\mathcal{L}_\text{proxy-based}$).}
	\begin{center}
		\begin{tabular}{|c|c|c|ccc|}
			\hline
			\multicolumn{1}{|c|}{$\lambda_\text{pps}$} & \multicolumn{1}{c|}{$\lambda_\text{pp}$} & \multicolumn{1}{c|}{$\lambda_\text{pns}$} & High-quality & IJB-B/C & TinyFace \\ 
			\hline
			$0$ & $150$ & $20$ & $93.59$ & $\bm{59.93}$ & $67.01$\\
			$5$ & $150$ & $20$ & $93.72$ & $59.40$ &$\bm{67.46}$\\
			$10$ & $150$ & $20$ & $\bm{93.89}$ & $58.68$ &$66.20$\\
			\hline
		\end{tabular}
		\\
	\end{center}
	\label{table:lambda_uamf}
\end{table}

\vfill

\end{document}